\documentclass[nohyperref]{article}

\usepackage{microtype}
\usepackage{graphicx}
\usepackage{subfigure}
\usepackage{booktabs}
\usepackage{hyperref}
\usepackage{multirow}

\usepackage[acronym,smallcaps,nowarn,section,nogroupskip,nonumberlist]{glossaries}
\glsdisablehyper{}

\usepackage{amsmath, amssymb, amsthm}
\usepackage{mathtools}
\theoremstyle{plain}

\theoremstyle{definition}

\theoremstyle{remark}

\newcommand{\calB}{{\mathcal{B}}}

\newcommand{\calD}{{\mathcal{D}}}

\newcommand{\calF}{{\mathcal{F}}}

\newcommand{\calH}{{\mathcal{H}}}

\newcommand{\calN}{{\mathcal{N}}}

\newcommand{\calS}{{\mathcal{S}}}
\newcommand{\calT}{{\mathcal{T}}}
\newcommand{\calU}{{\mathcal{U}}}

\newcommand{\bbR}{\mathbb{R}}

\newcommand{\bsd}{\boldsymbol{d}}

\newcommand{\bsf}{\boldsymbol{f}}

\newcommand{\bsp}{\boldsymbol{p}}
\newcommand{\bsq}{\boldsymbol{q}}
\newcommand{\bsr}{\boldsymbol{r}}
\newcommand{\bss}{\boldsymbol{s}}

\newcommand{\bsw}{\boldsymbol{w}}
\newcommand{\bsx}{\boldsymbol{x}}
\newcommand{\bsy}{\boldsymbol{y}}
\newcommand{\bsz}{\boldsymbol{z}}

\newcommand{\bvarepsilon}{{\boldsymbol{\varepsilon}}}

\newcommand{\btheta}{{\boldsymbol{\theta}}}

\DeclareMathOperator*{\argmax}{arg\,max}
\DeclareMathOperator*{\argmin}{arg\,min}

\newcommand{\kld}{D_{\mathrm{KL}}}
\newcommand{\tr}{^\top}

\def\[#1\]{\begin{align}#1\end{align}}
\newcommand{\norm}[1]{\left\lVert#1\right\rVert}

\newcommand{\spm}[1]{\scriptstyle{\pm#1}}

\PassOptionsToPackage{usenames,dvipsnames}{xcolor}

\usepackage[capitalize,noabbrev]{cleveref}

\usepackage[accepted]{icml2022}

\usepackage{tikz}
\usepackage{float}

\usetikzlibrary{tikzmark}

\newsavebox\CBox 
\def\textBF#1{\sbox\CBox{#1}\resizebox{\wd\CBox}{\ht\CBox}{\textbf{#1}}}

\icmltitlerunning{Improving Ensemble Distillation With Weight Averaging and Diversifying Perturbation}

\begin{document}

\twocolumn[
\icmltitle{Improving Ensemble Distillation With\\Weight Averaging and Diversifying Perturbation}

\begin{icmlauthorlist}
\icmlauthor{Giung Nam}{kaist}
\icmlauthor{Hyungi Lee}{kaist}
\icmlauthor{Byeongho Heo}{naver}
\icmlauthor{Juho Lee}{kaist,aitrics}
\end{icmlauthorlist}

\icmlaffiliation{kaist}{Korea Advanced Institute of Science and Technology, Daejeon, Korea}
\icmlaffiliation{aitrics}{AITRICS, Seoul, South Korea}
\icmlaffiliation{naver}{Naver, Korea}
\icmlcorrespondingauthor{Giung Nam}{giung@kaist.ac.kr}
\icmlcorrespondingauthor{Juho Lee}{juholee@kaist.ac.kr}

\icmlkeywords{Machine Learning, ICML}

\vskip 0.3in
]

\printAffiliationsAndNotice{}

\begin{abstract}
Ensembles of deep neural networks have demonstrated superior performance, but their heavy computational cost hinders applying them for resource-limited environments. It motivates distilling knowledge from the ensemble teacher into a smaller student network, and there are two important design choices for this ensemble distillation: 1) how to construct the student network, and 2) what data should be shown during training. In this paper, we propose a weight averaging technique where a student with multiple subnetworks is trained to absorb the functional diversity of ensemble teachers, but then those subnetworks are properly averaged for inference, giving a single student network with no additional inference cost. We also propose a perturbation strategy that seeks inputs from which the diversities of teachers can be better transferred to the student. Combining these two, our method significantly improves upon previous methods on various image classification tasks.
\end{abstract}

\section{Introduction}\label{main:sec:intro}
Deep Ensemble~\citep[DE;][]{lakshminarayanan2017simple} averages outputs of multiple models of the same architecture trained with the same data. Despite being simple to implement, DE achieves state-of-the-art performances for various tasks, serving as an oracle for many algorithms to compare against. However, the computational cost of DE scales linearly with the number of models involved in the ensemble, both for training and inference. Especially, the inference cost often becomes critical for a real-world scenario where both memory and time budget is limited.

Knowledge Distillation~\citep[KD;][]{hinton2015distilling} is a method to transfer knowledge from a large teacher network to a smaller student network. The heavy inference cost of DE thus naturally motivates applying KD to reduce it, where a DE is set as a teacher and a single neural network is introduced to be set as a student network. This task of distilling an ensemble teacher network (or distilling from multiple teachers if we treat each ensemble member as a teacher) is often called \emph{ensemble distillation} and has recently been studied actively~\citep{tran2020hydra,mariet2021distilling,nam2021diversity,ryabinin2021scaling,du2020agree}.

For a successful ensemble distillation, one should carefully choose the architecture for a student network. The most straightforward choice would be using a single neural network having the same architecture as the teacher, but this usually yields suboptimal results due to the limited flexibility of the student network. Another choice is to use a student network having \emph{subnetworks}, where the subnetworks share most of the parameters but have a small number of individual parameters, e.g., rank-one factors~\citep{wen2019batchensemble,mariet2021distilling} or multiple classification heads~\citep{tran2020hydra}. The ensemble distillation with subnetworks are reported to improve performance upon vanilla ensemble distillation~\citep{mariet2021distilling,tran2020hydra,nam2021diversity}, but they usually require additional computational costs for inference. For instance, the inference cost of a student network having subnetworks defined with rank-one factors scales linearly to the number of subnetworks.

Another important choice for an ensemble distillation algorithm is the training data perturbation strategy. Recently, \citet{nam2021diversity} studied the importance of diversities in ensemble distillation. When ensemble teachers achieve near-zero train error, the outputs of ensemble teachers for a training data point would be nearly identical, so the ensemble distillation with normal training data would not effectively transfer diversities of the ensemble teachers to students. To this end, \citet{nam2021diversity} proposed to perturb training data to output-diversifying directions~\citep{tashiro2020diversity} and use them for distillation. While this indeed improves performance, the perturbation strategy proposed in \citet{nam2021diversity} only considers teachers without considering how a student would react to such perturbed data. Student networks are typically less flexible than teacher networks, so a perturbation increasing diversities of teachers may act in an unexpected way when applied to students. Hence, we were motivated to consider both student diversities and teacher diversities into account when designing a perturbation strategy. 

In this paper, we propose a novel ensemble distillation method that resolves both of the above-mentioned limitations. Our first contribution is the new way of constructing a student network; we propose to distill with a student with multiple subnetworks during training, but average the subnetwork weights later for inference to get a single student network as a result. As a prototype, we apply this idea to BatchEnsemble~\citep[BE;][]{wen2019batchensemble}, where the subnetworks are differentiated with multiplicative rank-one factors. Specifically, for such a student network, we propose a training scheme that encourages the subnetwork parameters to stay within the same low-loss region so that the averaged prediction does not degrade the performance while maximally absorbing the diversities transferred from teachers. We show that under a representative training scheme on image classification, averaging those rank-one factors does not degrade predictive performance. The second contribution is a novel perturbation strategy improving upon the one proposed in \citet{nam2021diversity}. Unlike the previous method, our perturbation method considers both students and teachers. More specifically, we find the ``weak points’’ of the student networks by seeking inputs on which the student subnetworks agree with each other (low diversity), and at the same time, find the inputs on which the teachers disagree (high diversity). With this perturbation considering both students and teachers, our method effectively transfers diversities of teachers to students. To demonstrate the effectiveness of our method, we compare ours with previous methods on various image classification benchmarks. We find that ours achieve significantly improved predictive accuracy and uncertainty calibration results without increasing inference cost.

\section{Backgrounds}
\subsection{Setup}
The problem we address in this paper is the $K$-way classification problem; a neural network $\calF:\bbR^D \rightarrow \bbR^K$ takes $D$-dimensional inputs $\bsx$ (i.e., images) and makes predictions about corresponding outputs $y$ (i.e., class label) with $K$-dimensional logits $\calF(\bsx)$. We denote the output probabilities of the model $\calF$ for a given input $\bsx$ as
\[
\bsp_{\calF}^{(k)}(\bsx ; \tau) = \frac{
    \exp{\left( \calF^{(k)}(\bsx) / \tau \right)}
}{
    \sum_{j=1}^{K} \exp{\left( \calF^{(k)}(\bsx) / \tau \right)}
},
\]
for $k=1,...,K$. Here, we introduce a single scale parameter $\tau>0$ for temperature scaling which will be used both for training and evaluation procedures.

\subsection{Ensemble Distillation}
Let $\{\calT_1,...,\calT_M\}$ be a set of pre-trained teachers, and $\calS_\btheta$ be a student. In a vanilla ensemble distillation, using the knowledge distillation~\citep[KD;][]{hinton2015distilling} framework, the student tries to mimic the probabilistic outputs of an ensemble of teachers under the given temperature $\tau$ by minimizing the averaged KD loss, which is equivalent to minimizing
\[\label{eq:ekd}
\tau^2 \calH \left[
    \frac{1}{M} \sum_{m=1}^{M} \bsp_{\calT_m}(\bsx;\tau), \bsp_{\calS_\btheta}(\bsx;\tau)
\right],
\]
where $\calH[\cdot,\;\cdot]$ computes the cross-entropy between two probability vectors. Note that this vanilla approach minimizes the discrepancy between the student predictions and the \emph{mean} predictions of the ensemble teachers. As a result, the diversities among ensemble teachers are removed by mean operation and hardly transferred to the student.

\subsection{BatchEnsemble and one-to-one distillation}
BatchEnsemble~\citep[BE;][]{wen2019batchensemble} is a parameter-efficient way to ensemble deep neural networks; each member of the ensemble is constructed in the low-rank subspace with rank-one factors, instead of the full parameter space. With a slight abuse of notation, while a DE would have full set of parameters $\{\btheta_1, \dots, \btheta_M\}$, BE introduces a shared parameter $\btheta$ and a set of rank-one matrices $\{\bsr_1\bss_1\tr, \dots, \bsr_m\bss_m\tr\}$, and construct $m^\text{th}$ subnetwork parameter as $\btheta \circ \bsr_m\bss_m\tr$, where $\circ$ denotes the Hadamard product. Based on BE, \citet{mariet2021distilling} proposed a one-to-one ensemble distillation scheme, where each BE subnetwork is trying to mimic single ensemble member in one-to-one fashion. Instead of learning the mean prediction of the teachers, the one-to-one distillation minimizes
\[
\sum_{m=1}^M \tau^2 \calH\left[
    \bsp_{\calT_m}(\bsx;\tau), \bsp_{\calS_{\btheta\circ \bsr_m\bss_m\tr}}(\bsx;\tau)
\right].
\]
That is, each subnetwork copies a member from the ensembles in a one-to-one way. The training procedure for BE ensemble distillation is summarized in \cref{algorithm_EKD_BE}.

\begin{algorithm}[t]
\caption{Ensemble distillation with BE}
\label{algorithm_EKD_BE}
\begin{algorithmic}[1]
\REQUIRE{Temperature $\tau$, learning rate $\eta$.}
\color{black}
\WHILE{not converged}
    \STATE Sample an input $\bsx$ from the train split.
    \FOR{$m=1,...,M$}
        \STATE Compute loss for the $m^\text{th}$ subnetwork:
        
        \quad$\ell_m \gets \tau^2 \calH\left[ \bsp_{\calT_m}(\bsx;\tau), \bsp_{\calS_{\btheta\circ(\bsr_m \bss_m\tr)}}(\bsx;\tau) \right]$.
        \STATE Update rank-one factors:
        
        \quad$\bsr_m \gets \bsr_m - \eta \nabla_{\bsr_m} \ell_m$.
        
        \quad$\bss_m \gets \bss_m - \eta \nabla_{\bss_m} \ell_m$.
    \ENDFOR
    \STATE Update shared parameters:
    
    \quad$\btheta \gets \btheta - \eta \frac{1}{M} \sum_{m=1}^{M} \nabla_{\btheta} \ell_m$.
\ENDWHILE
\end{algorithmic}
\end{algorithm}

\subsection{Ensemble distillation with diversifying perturbation}\label{main:subsec:perturb}
When the ensemble teachers are flexible enough to achieve zero-train error, their responses to a training input would be nearly identical, so a student network distilled from them would not be exposed to the diversities of the teachers. To resolve this, \citet{nam2021diversity} proposes to perturb training inputs with the Output Diversified Sampling~\citep[ODS;][]{tashiro2020diversity} that encourages ensemble teachers disagree with each other. The ODS for an input $\bsx$ is computed as
\[
\bvarepsilon_{\text{ODS}} &\propto \nabla_{\bsx}\left( \bsw\tr \bsp_{\calT_m}(\bsx;\tau) \right),
\]
where $\bsw$ denotes a random guidance vector sampled from the $K$-dimensional uniform distribution with zero means. Intuitively, the ODS perturbation seeks the direction in the input space to make the output follow the random guidance vector $\bsw$, and thus drives ensemble members to produce diverse outputs~\footnote{Actually, we need a transferability assumption to fully justify this argument. For more detail, please refer to \citet{nam2021diversity}.}. \citet{nam2021diversity} demonstrated that a BE student trained with ODS perturbation shows significantly improved performance. They also proposed an improved version of ODS perturbation called ConfODS, where the ODS perturbations are scaled by the confidence of teacher predictions.

\section{Improved Ensemble Distillation}
\subsection{LatentBE : a weight averaged BE student}
Utilizing the subnetwork structure, a BE student can capture the diversities of ensemble teachers, but this comes at a cost of increased inference time. To get an output from a BE network, one should execute forward passes $M$ times from scratch since the different rank-one subnetworks do not share hidden layer responses. Hence, although BE significantly reduces the number of parameters compared to DE, its computation cost for inference is identical to that of DE. This is definitely undesirable, especially for distillation where we want a cheap student network applicable for real-world applications. In this section, we propose a novel ensemble distillation framework to circumvent this limitation, where a student network has the same inference cost as a single neural network yet maintains the flexibility to well absorb the diversity of teachers.

Based on the ensemble distillation with BE, we propose a novel framework entitled LatentBE, in a sense that the rank-one factors defining BE are averaged out for inference just as the latent variables are marginalized out for probabilistic inference. Specifically, we employ a BE student having multiple subnetworks, and do one-to-one ensemble distillation similarly to \cref{algorithm_EKD_BE}, but after training, compute the \emph{weight average} of the rank-one factors to construct a single student network with parameter
\[
\btheta \circ \bigg(\frac{1}{M}\sum_{m=1}^M \bsr_m\bss_m\tr\bigg).
\]
After this weight averaging, the inference cost remains the same as that of a single neural network. The idea of weight averaging was first considered in Stochastic Weight Averaging~\citep[SWA;][]{izmailov2018averaging}, where the parameters collected from a single learning trajectory is averaged to construct a better generalizing model. The key observation in SWA is that due to the choice of the specific learning rate scheduling, the parameters in a learning trajectory remain in a wide low-loss region in the loss surface, so averaging them leads to a single robust model.  Our LatentBE shares some spirits with SWA but has crucial differences: 1) LatentBE utilizes the diverse subnetworks from the guidance of ensemble teachers, which brings diversity and performance gain of ensemble distillation to \emph{weight average}, and 2) instead of special learning rates schedule of SWA, LatentBE enables \emph{weight average} with the rank-one factors of BE.

\begin{algorithm}[t]
\caption{Ensemble distillation with LatentBE + diversifying perturbation}
\label{algorithm_EKD_LatentBE}
\begin{algorithmic}[1]
\REQUIRE{Temperature $\tau$, learning rate $\eta$, {\color{RoyalBlue} rank-one weight decay parameter $\lambda$, perturbation step size $\gamma$.}}
{\color{RoyalBlue}
\STATE Initialize rank-one factors to ones:

\quad$\bsr_m \gets \boldsymbol{1}$ and $\bss_m \gets \boldsymbol{1}$ for $m=1,...,M$.
}
\WHILE{not converged}
    \STATE Sample an input $\bsx$ from the train split.
    {\color{RoyalBlue}
    \STATE Sample indices $i, j$ uniformly from $\{1, \dots, M\}$.
    \STATE Perturb the input w.r.t. teacher and student:
    
    \quad$\tilde{\bsx} \gets \bsx + \gamma (\widehat{\operatorname{TDiv}}(\bsx) - \widehat{\operatorname{SDiv}}(\bsx)).$
    }
    \FOR{$m=1,...,M$}
        \STATE Compute loss for the $m^\text{th}$ subnetwork:
        
        \quad$\ell_m \gets \tau^2 \calH\left[ \bsp_{\calT_m}({\color{RoyalBlue}\tilde{\bsx}};\tau), \bsp_{\calS_{\btheta\circ(\bsr_m \bss_m\tr)}}({\color{RoyalBlue}\tilde{\bsx}};\tau) \right]$
        \STATE Update rank-one factors:
        
        \quad$\bsr_m \gets \bsr_m - \eta \nabla_{\bsr_m} \ell_m - {\color{RoyalBlue}\eta\lambda(\bsr_m-\boldsymbol{1})}.$
        
        \quad$\bss_m \gets \bss_m - \eta \nabla_{\bss_m} \ell_m - {\color{RoyalBlue}\eta\lambda(\bss_m - \boldsymbol{1})}.$
    \ENDFOR
    \STATE Update shared parameters:
    
    \quad$\btheta \gets \btheta - \eta \frac{1}{M} \sum_{m=1}^{M} \nabla_{\btheta} \ell_m$.
\ENDWHILE
{\color{RoyalBlue}
\STATE Return the averaged parameter:

\quad$\btheta \gets \btheta \circ \frac{1}{M}\sum_{m=1}^M \bsr_m\bss_m\tr$.
}
\end{algorithmic}
\end{algorithm}

The key for making LatentBE successful is, as in SWA, to keep the subnetwork parameters stay in the same low-loss region. For this, 1) we initialized all the rank-one factors $\{\bsr_m, \bss_m\}_{m=1}^M$ to be one vectors, and 2) set Gaussian prior with one vector mean to those rank-one factors. By doing these, all the rank-one factors start from a similar location and gradually split into individual factors, but they do not deviate too much from each other due to the weight-decay effect driven by the prior. Also, other than these two, we do not require careful learning rate scheduling as in SWA, and this is presumably due to the fact that we are only differentiating rank-one factors for subnetworks with a large body of shared parameters while SWA averages entire parameters. 

\begin{figure}[t]
    \centering
    \subfigure[BE-2]{\frame{\includegraphics[width=0.48\linewidth]{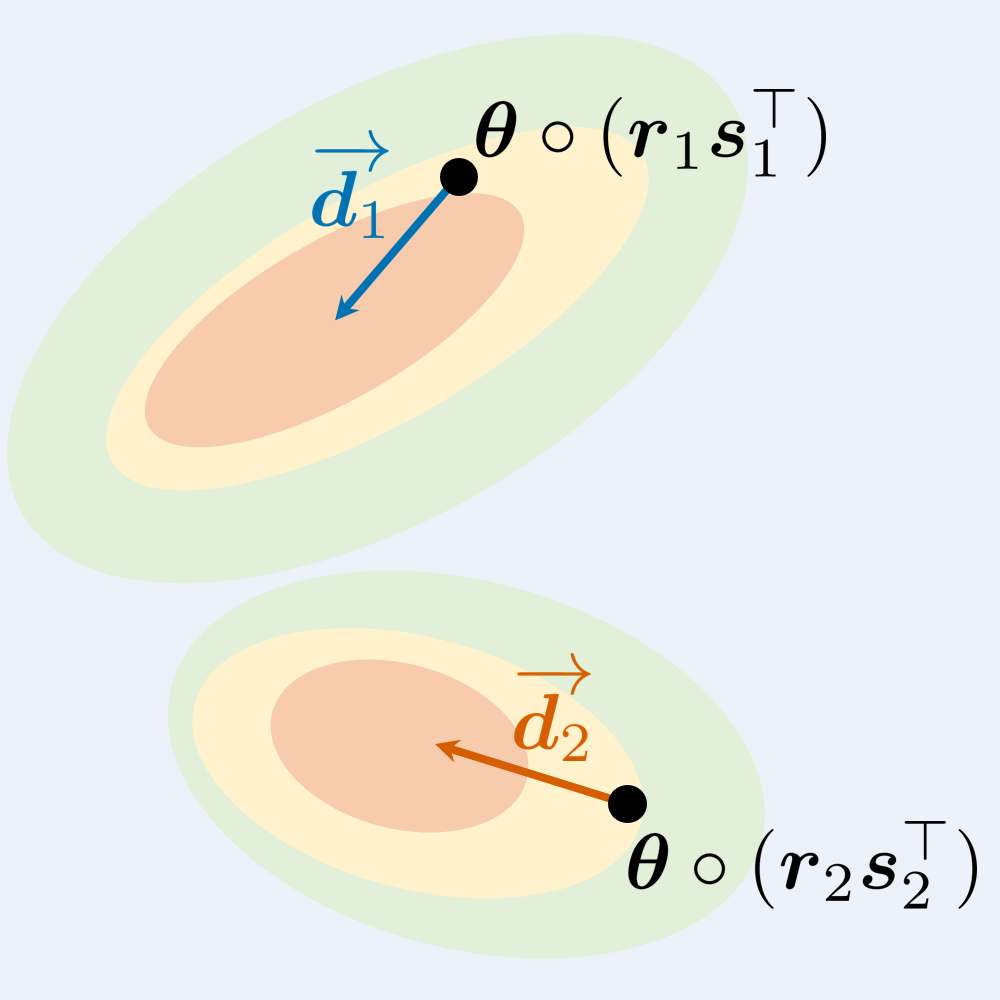}}}\hfill
    \subfigure[LatentBE-2]{\frame{\includegraphics[width=0.48\linewidth]{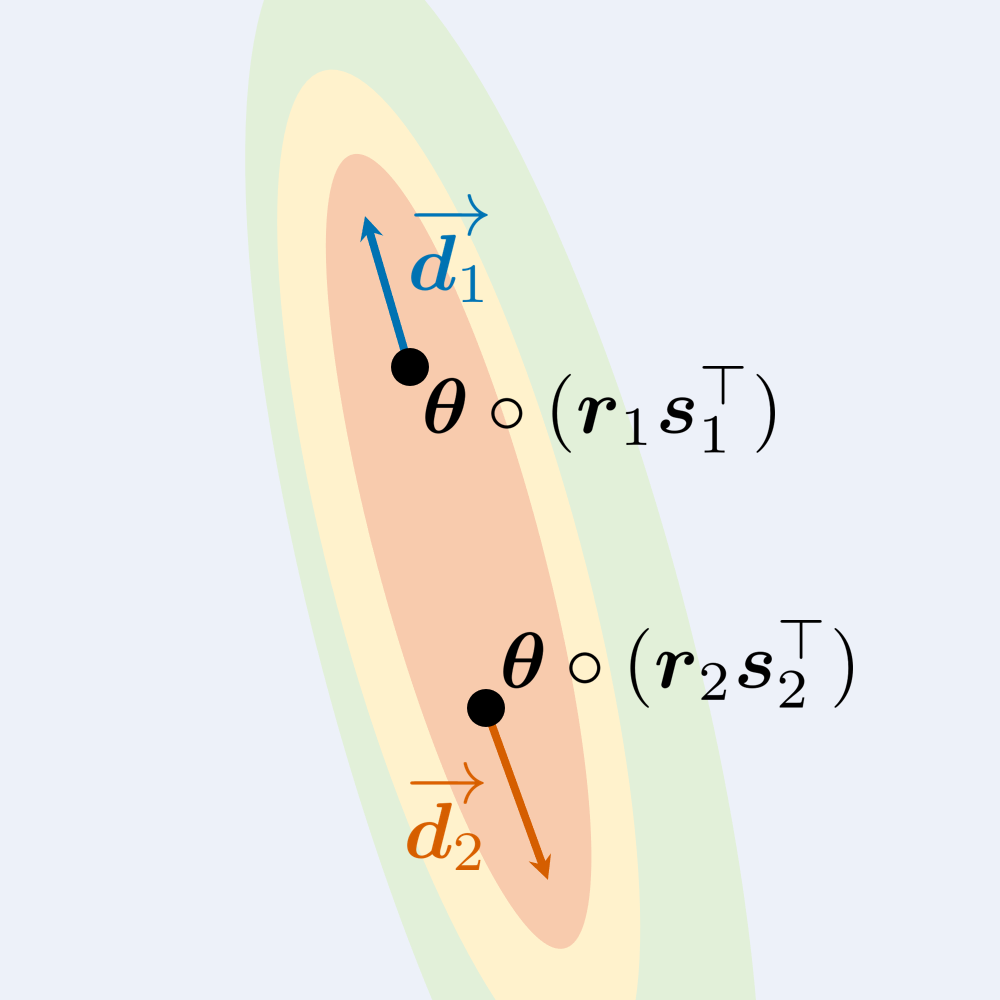}}}
    \caption{A schematic diagram depicting BE and LatentBE. Here, $\overrightarrow{\bsd_1}$ and $\overrightarrow{\bsd_2}$ denote learning directions obtained from two different teachers. For the BE, (a) two subnetwork parameters \emph{explore} different modes with low-rank subspaces, while for the LatentBE, (b) two subnetwork parameters \emph{expand} a low-rank subspace.}
    \label{fig:motivating}
\end{figure}

The rank-one factors for BE and LatentBE play different roles in ensemble distillation. BE aims to find a subspace for each rank-one factor and spread those subspaces to different modes. On the other hand, LatentBE seeks for a flat minima in which all the subspaces defined by the rank-one factors are embedded, and then the rank-one factors are trained to ``stretch" the subspace area by letting the subnetworks follow different teacher directions~(\cref{fig:motivating}).

\subsection{A better perturbation strategy}\label{main:sec:perturbation}
As we discussed earlier, \citet{nam2021diversity} utilizes perturbation strategies, ODS and ConfODS, to improve diversity transfer in ensemble distillation. It is an innovative approach for ensemble distillation but does not consider the student networks, especially their diversities. Since there have been several works controlling teacher networks based on the status of student networks for KD~\cite{jin2019knowledge,mirzadeh2020improved}, we further conjecture that the ensemble distillation can also be improved with a perturbation strategy considering both teachers and students. Thus, we propose a novel perturbation scheme that considers \emph{both} student and teacher diversities for more effective diversity transfer.

For a given input $\bsx$, we measure the functional diversity of the ensemble of $\{\calF_1,...,\calF_M\}$ by averaging pairwise KL-divergence between probabilistic outputs from different members,
\[\label{eq:div}
\operatorname{Div} \left(
    \left\{ \calF_m \right\}_{m=1}^{M}, \bsx
\right) = \frac{\sum_{i=1}^{M} \sum_{j=1}^{M} D_{ij}(\bsx)}{M(M-1)},
\]
where $D_{ij}(\bsx) = \kld ( \bsp_{\calF_i}(\bsx)\;\Vert\;\bsp_{\calF_j}(\bsx) )$. From this, we denote the student and teacher diversities as
\[
\operatorname{SDiv}(\bsx) &\coloneqq \operatorname{Div} \left(
    \left\{ \calS_{m} \right\}_{m=1}^{M}, \bsx
\right), \label{eq:sdiv}\\
\operatorname{TDiv}(\bsx) &\coloneqq \operatorname{Div} \left(
    \left\{ \calT_{m} \right\}_{m=1}^{M}, \bsx
\right) \label{eq:tdiv},
\]
where $\calS_m \coloneqq \calS_{\btheta\circ \bsr_m\bss_m\tr}$. We suggest perturbing input to the direction that \emph{minimizes} the student diversity while the teacher diversity is \emph{maximized},
\[
\bvarepsilon \propto \nabla_{\bsx} \left(
    \operatorname{TDiv}(\bsx) - \operatorname{SDiv}(\bsx)
\right).
\]
The intuition behind this perturbation is as follows: ideally, we want the student subnetworks to learn the diverse outputs of ensemble teachers almost everywhere in the input space. Hence, we first introduce negative student diversity term ``$- \operatorname{SDiv}(\bsx)$'' to find a point that student subnetworks have low diversities. At the same time, as we originally intended, we use teacher diversity term ``$\operatorname{TDiv}(\bsx)$'' to improve diversity transfer of ensemble teachers. The combined perturbation thus finds an input point that maximizes the diversity gap between teachers and student subnetworks and gives a strong learning signal to correct it.

In practice, exactly computing \eqref{eq:sdiv} and \eqref{eq:tdiv} would be costly, so we use stochastic approximations of them where the student and teacher diversities are computed for a randomly selected pair. That is, we pick $i, j \sim \{1, \dots, M\}$ uniformly and compute
\[
\widehat{\operatorname{TDiv}}(\bsx) &\coloneqq \kld(
    \bsp_{\calT_i}(\bsx)\;||\;\bsp_{\calT_j}(\bsx)
), \\
\widehat{\operatorname{SDiv}}(\bsx) &\coloneqq \kld(
    \bsp_{\calS_{i}}(\bsx)\;||\;\bsp_{\calS_{j}}(\bsx)
),
\]
and define the perturbation as
\[
\hat{\bvarepsilon} \propto \nabla_{\bsx} \left(
    \widehat{\operatorname{TDiv}}(\bsx) - \widehat{\operatorname{SDiv}}(\bsx)
\right).
\]
We also found that blocking the gradient flow through one of the teachers or students stabilizes the training,
\[
\widehat{\operatorname{TDiv}}(\bsx) &\coloneqq \kld(
    \texttt{sg}(\bsp_{\calT_i}(\bsx))\;\Vert\;\bsp_{\calT_j}(\bsx)
), \\
\widehat{\operatorname{SDiv}}(\bsx) &\coloneqq \kld(
    \texttt{sg}(\bsp_{\calS_{i}}(\bsx))\;\Vert\;\bsp_{\calS_{j}}(\bsx)
),
\]
where $\texttt{sg}(\cdot)$ denotes the $\texttt{stop\_grad}$ operation, for example, $\texttt{.detach()}$ in PyTorch library.

One thing to note here is that we are directly measuring divergences between teachers or student subnetworks to get perturbations, unlike the ODS-based perturbation proposed in \citet{nam2021diversity}. An ODS computed from a specific network, in principle, does not guarantee the output diversification of other networks. Hence, the ODS perturbation computed from a single teacher, as suggested in \citet{nam2021diversity}, does not guarantee the diversities among teacher outputs. \citet{nam2021diversity} argued that this issue can be circumvented by assuming \emph{transferability} of teacher networks, where we assume that the gradients of ensemble teachers are similar to each other, so an ODS computed from a specific teacher generalizes to the other teachers, driving overall diversities among teacher outputs as a result. However, the transferability does not always hold for which an ODS perturbation fails to properly bring diversities. On the other hand, our perturbation directly minimizing or maximizing KL divergences does not require transferability of gradients.

\subsection{Improved ensemble distillation algorithm}
Our final ensemble distillation algorithm combines two ingredients discussed so far; LatentBE and novel diversifying perturbation. The only overhead during training is the procedure of computing the diversifying perturbation which requires additional forward and backward passes through teacher networks. Our algorithm is summarized in \cref{algorithm_EKD_LatentBE}, with {\color{RoyalBlue} highlights} on the part different from the vanilla one-to-one ensemble distillation with BE.

\section{Related Works}
\paragraph{Ensembles}
Recent works have shown that an ensemble of deep neural networks can achieve superior performance both in terms of prediction accuracy and uncertainty estimation~\citep{lakshminarayanan2017simple,ovadia2019trust}. The power of the ensemble comes from \textit{the diversity} among ensemble members, and there have been several works to enhance it, e.g., constructing ensembles with varying hyperparameters~\citep{wenzel2020hyperparameter}, or architectures~\citep{zaidi2021neural}, or reducing conditional redundancy~\citep{rame2021dice}, or introducing kernelized repulsion~\citep{d2021repulsive}.

\paragraph{Ensemble distillation}
The seminal work of~\citet{hinton2015distilling} has already shown the effectiveness of the ensemble distillation. It can be further enhanced by considering the diversity inside the ensemble teacher, e.g., dynamically assign weights to teachers~\citep{du2020agree}, or treating predictions from teachers as a set of samples from an implicit distribution~\citep{malinin2019ensemble,ryabinin2021scaling}, or amplifying the diversity via input perturbations~\citep{nam2021diversity}. Besides, several existing approaches propose to use a student having subnetworks which can represent the diversity in predictions~\citep{tran2020hydra,mariet2021distilling,nam2021diversity}. However, their resulting students have additional costs for inference and defeat their ends that reduce the computational cost of the ensemble.

\section{Experiments}
In this section, we present the experimental results on image classification benchmarks including CIFAR-10, CIFAR-100~\citep{krizhevsky2009learning}, TinyImageNet, and ImageNet-1k~\citep{russakovsky2015imagenet}. Through the experiments, we empirically validate the following questions:
\begin{itemize}
    \item How does the subspace discovered by LatentBE look like? - \cref{main:subsec:latentbe_subspace}.
    \item How does the proposed perturbation strategy affect the training of LatentBE? - \cref{main:subsec:diversification}.
    \item Does our ensemble distillation algorithm improves performance both in terms of predictive accuracy and uncertainty calibration? - \cref{main:subsec:classification}.
\end{itemize}
Please refer to \cref{app:experimental_details} for the training details including data augmentation, learning rate schedules, and other hyperparameter settings.

\begin{figure*}[t]
    \centering
    \includegraphics[width=0.32\linewidth]{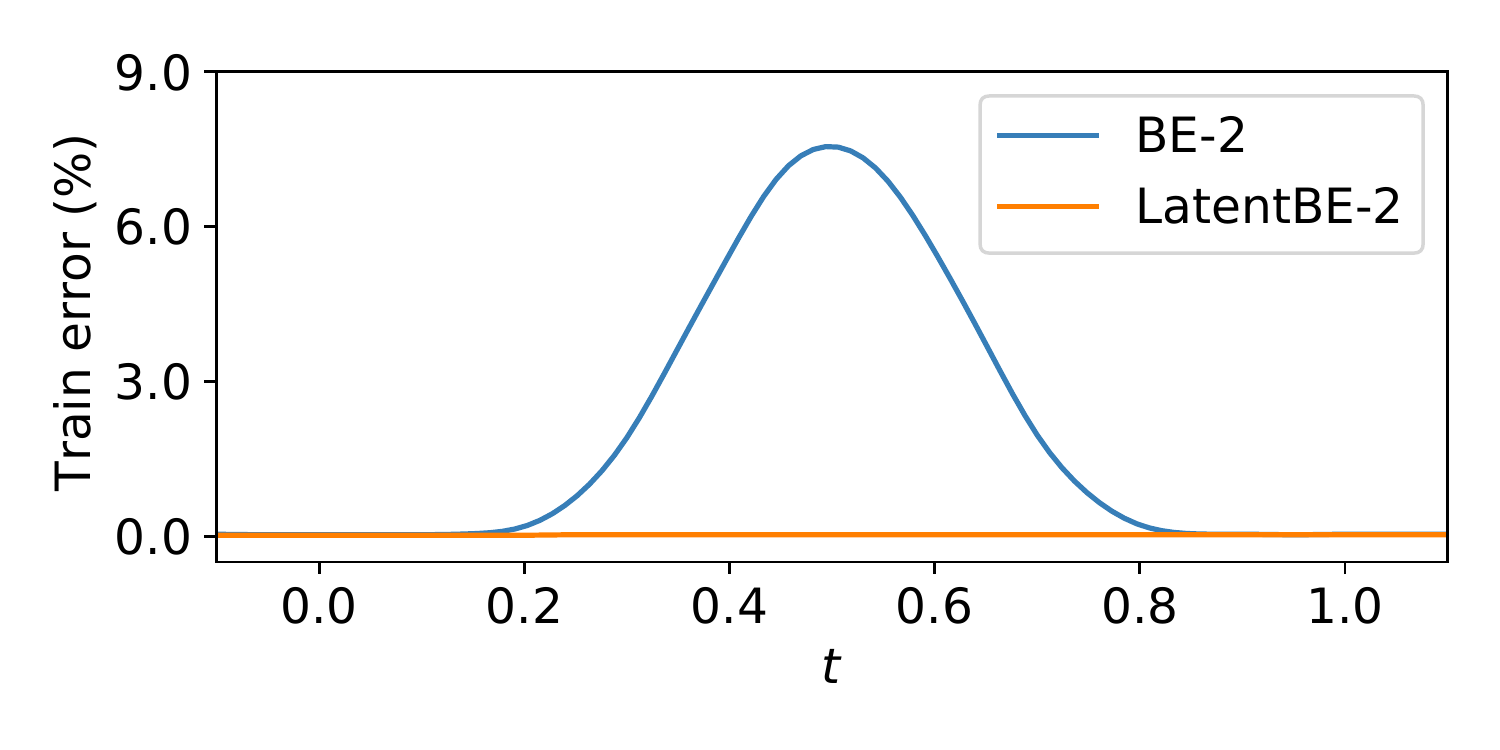}\hfill
    \includegraphics[width=0.32\linewidth]{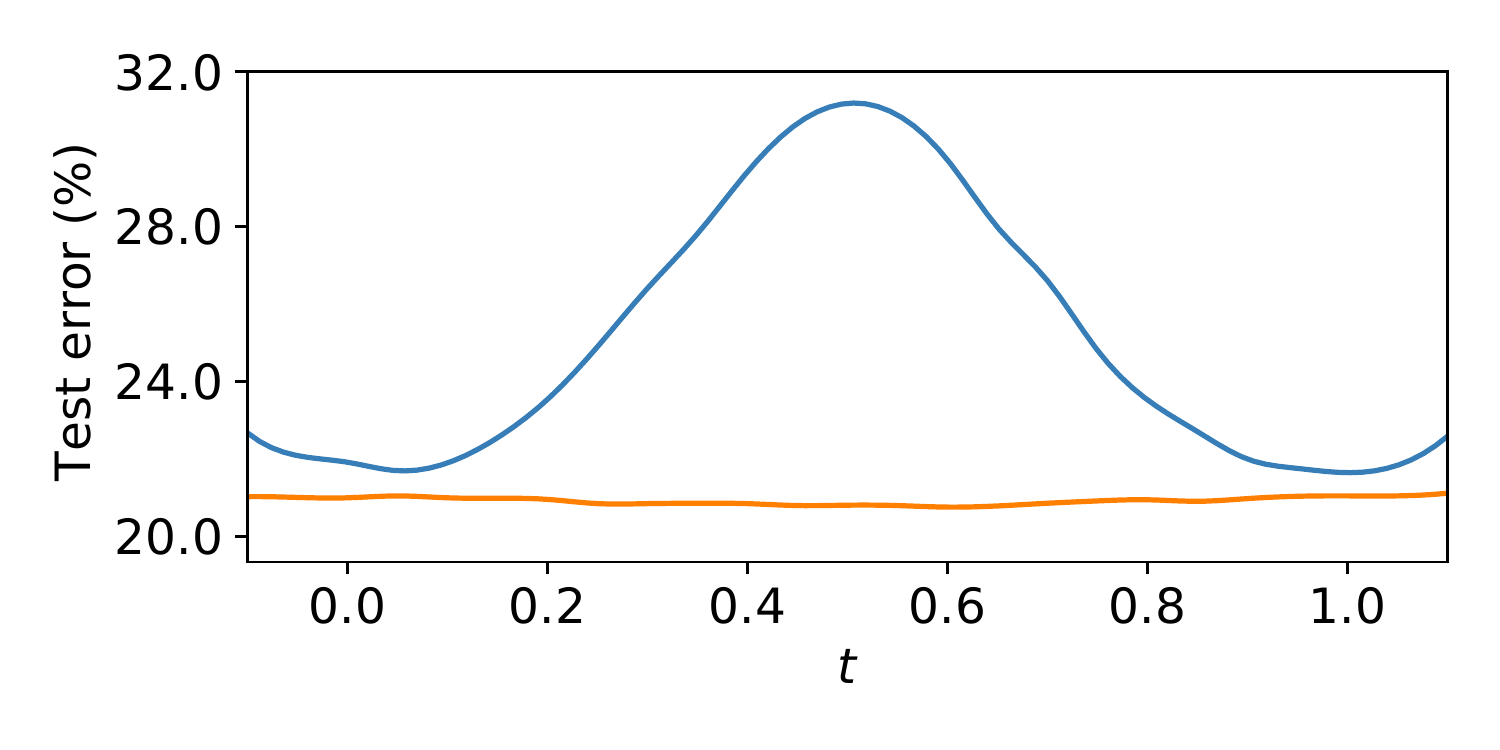}\hfill
    \includegraphics[width=0.32\linewidth]{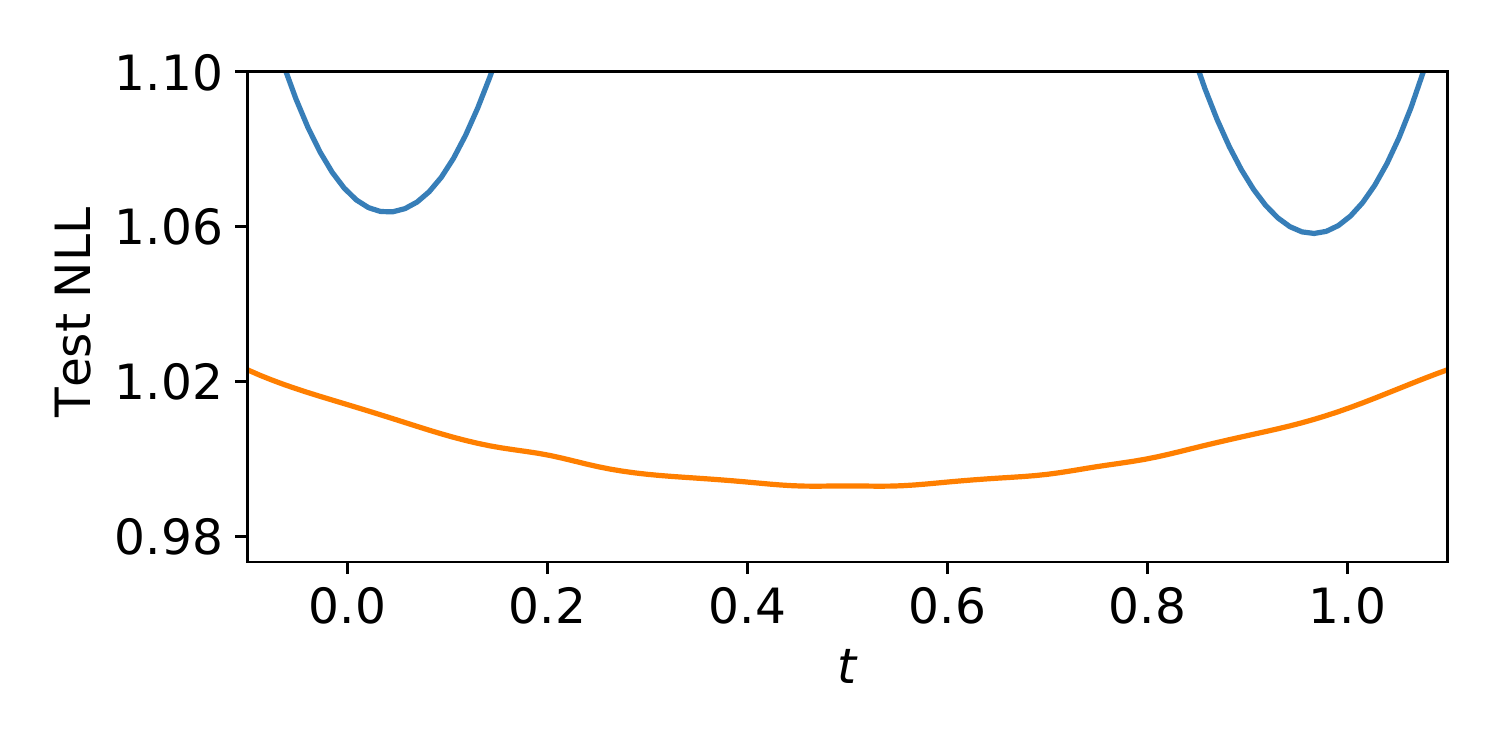}\vspace*{-5mm}
    \caption{Train erros (left), test errors (middle), and test negative log-likelihood along the line subspace (right) passing through two different subnetwork parameters. $t$ denotes the position on the line as defined in~\cref{eq:line}. BE-2 and LatentBE-2 students are distilled from the DE-2 teacher for WRN28x4 on CIFAR-100 \emph{without} any input perturbations.}
    \label{fig:subspace}
    \vspace{1em}
    \centering
    \includegraphics[width=0.32\linewidth]{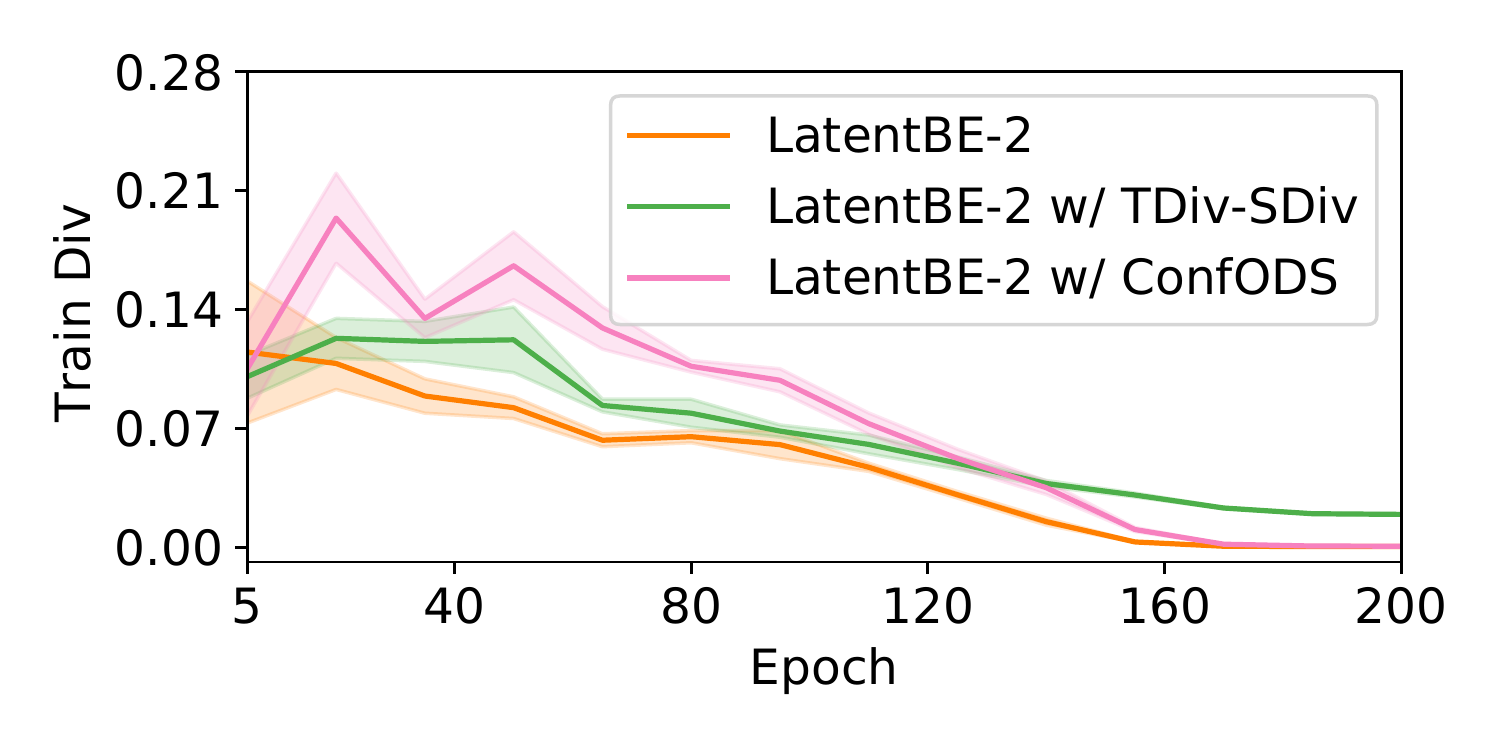}\hfill
    \includegraphics[width=0.32\linewidth]{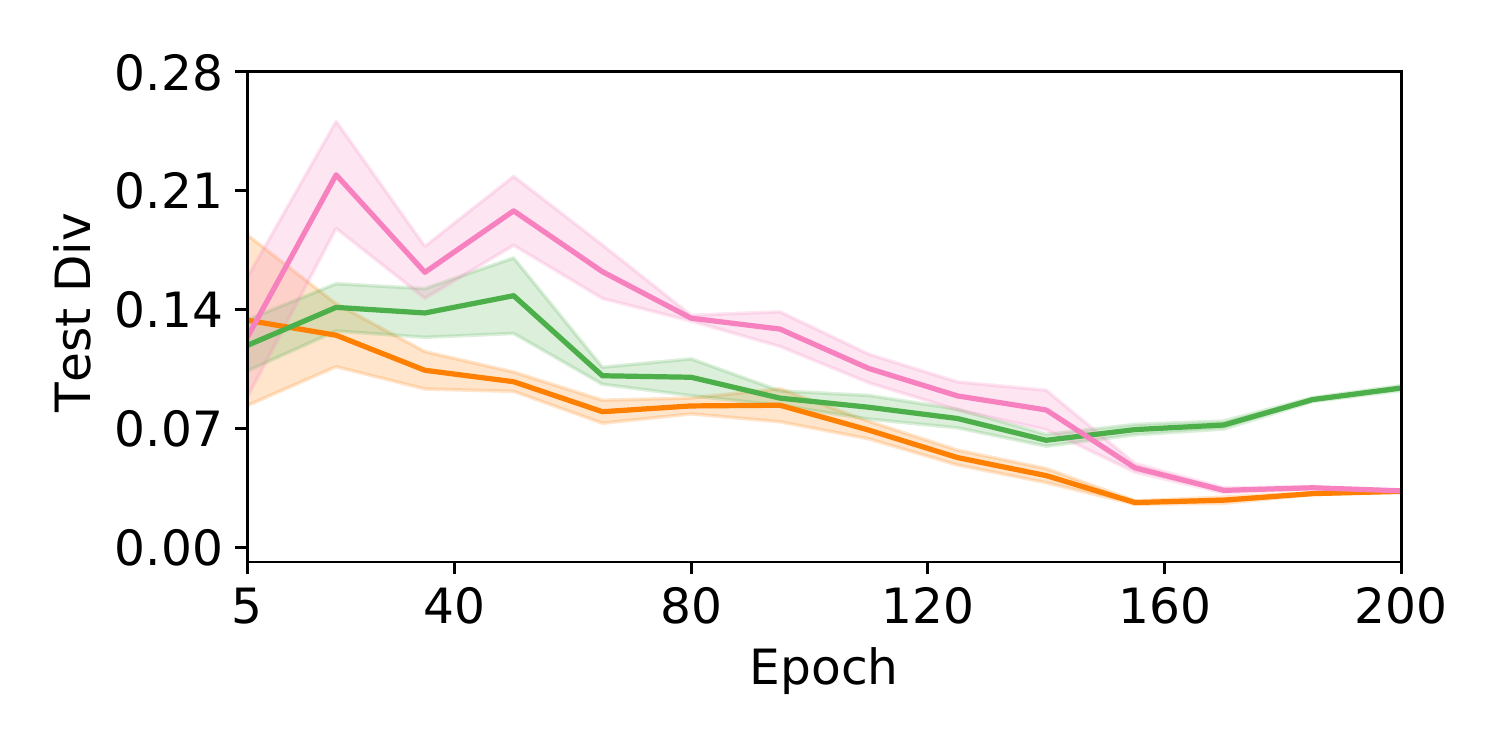}\hfill
    \includegraphics[width=0.32\linewidth]{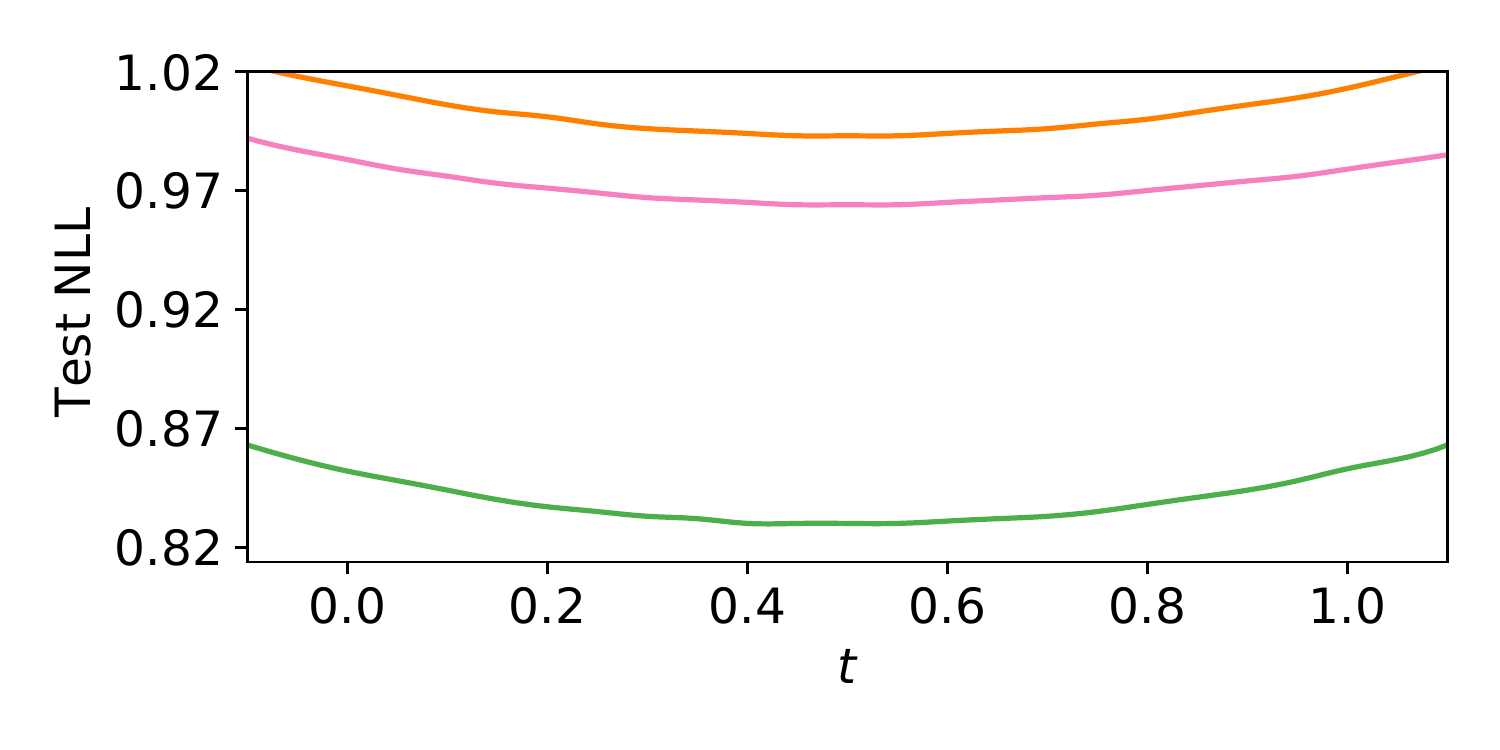}\vspace*{-5mm}
    \caption{Function diversity in predictions from two subnetwork parameters of the LatentBE-2 student on the train (left) and test (middle) data measured for a training run. We also visualize the resulting student's test NLL (right). LatentBE-2 students are distilled from the DE-2 teacher for WRN28x4 on CIFAR-100 \emph{with} the stated input perturbations.}
    \label{fig:endpoints}
\end{figure*}

\subsection{Subspaces of LatentBE}\label{main:subsec:latentbe_subspace}
In order to investigate the subspaces defined by the subnetwork parameters $\{\btheta\circ(\bsr_m\bss_m\tr)\}_{m=1}^{M}$, we first consider the case of $M=2$ models where the subspace forms a simple line. More precisely, we parameterize the line passing through two subnetwork parameters as $\left\{ \btheta_t\;|\;t\in\bbR \right\}$, where
\[\label{eq:line}
\btheta_t = (1-t) \left( \btheta\circ(\bsr_1\bss_1\tr) \right) + t \left( \btheta\circ(\bsr_2\bss_2\tr) \right).
\]
\cref{fig:subspace} shows how prediction error and negative log-likelihood vary along the line subspace. The main difference between BE and LatentBE students is the presence of \emph{a loss barrier}  between two end-points. The LatentBE student does not have a barrier while the BE student does, as we have depicted in \cref{fig:motivating}. This difference enables the weight averaging of subnetwork parameters for LatentBE. Notably, as shown in~(\cref{fig:subspace}, right), the averaged parameter effectively improves negative log-likelihood on the test data, which is consistent with the findings \citet{izmailov2018averaging} and recent study on the neural network subspaces~\citep{wortsman2021learning}. These improvements in performance, as we will show later in~\cref{main:subsec:classification}, confirm the validity of our weight averaging strategy for ensemble distillation.

\subsection{Diversification effects of perturbations}\label{main:subsec:diversification}
Using on the LatentBE-2 model discussed on the previous section, \cref{fig:endpoints} shows the functional diversity (defined in~\cref{eq:div}) between two end-points, when distilled with and without ConfODS~\citep{nam2021diversity} and ours (i.e., TDiv-SDiv). As one can see from the figures, ours better diversifies the subnetworks while maintaining lower test NLL values.

\begin{figure}
    \centering
    \includegraphics[width=0.95\linewidth]{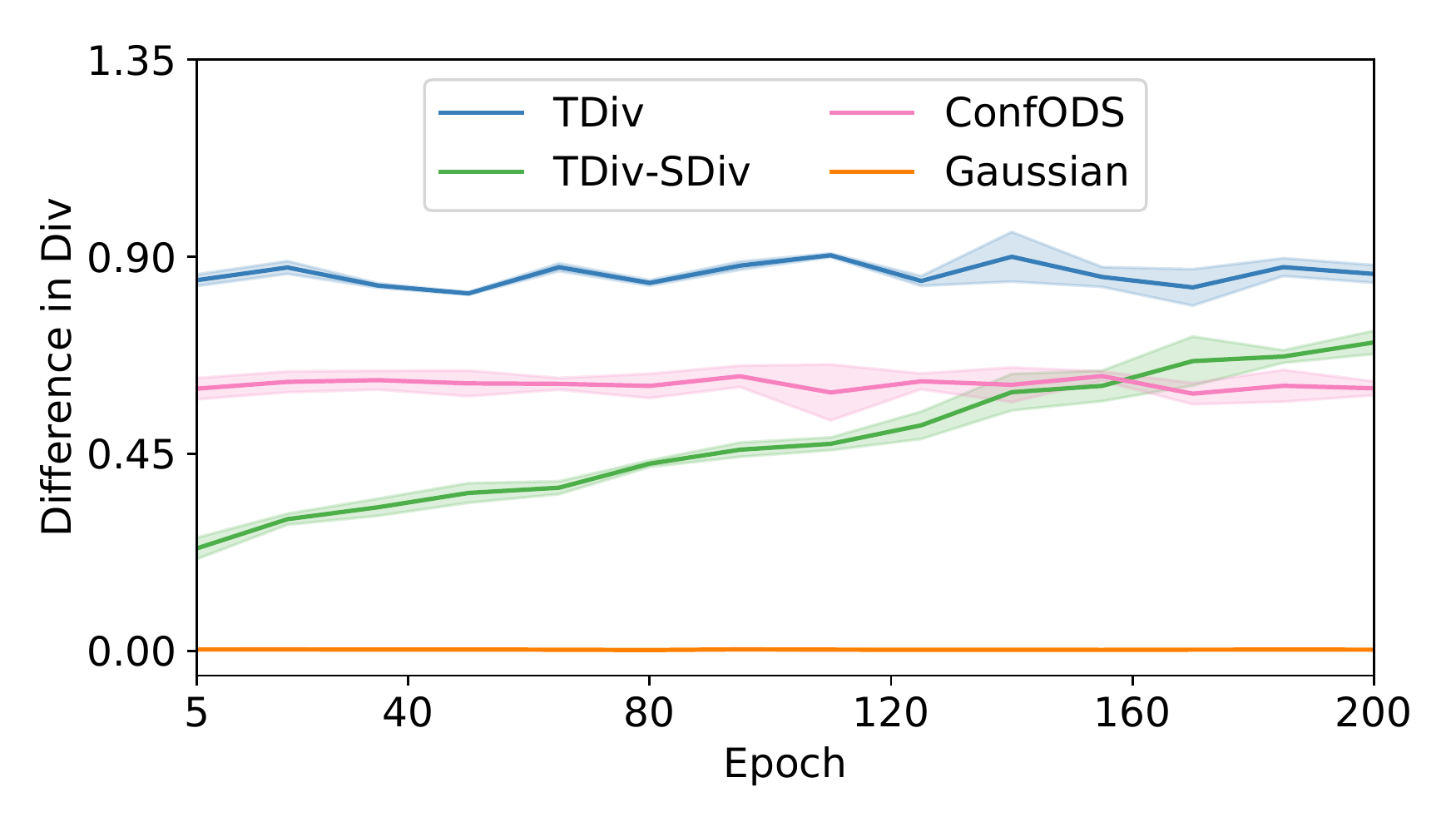}
    \includegraphics[width=0.95\linewidth]{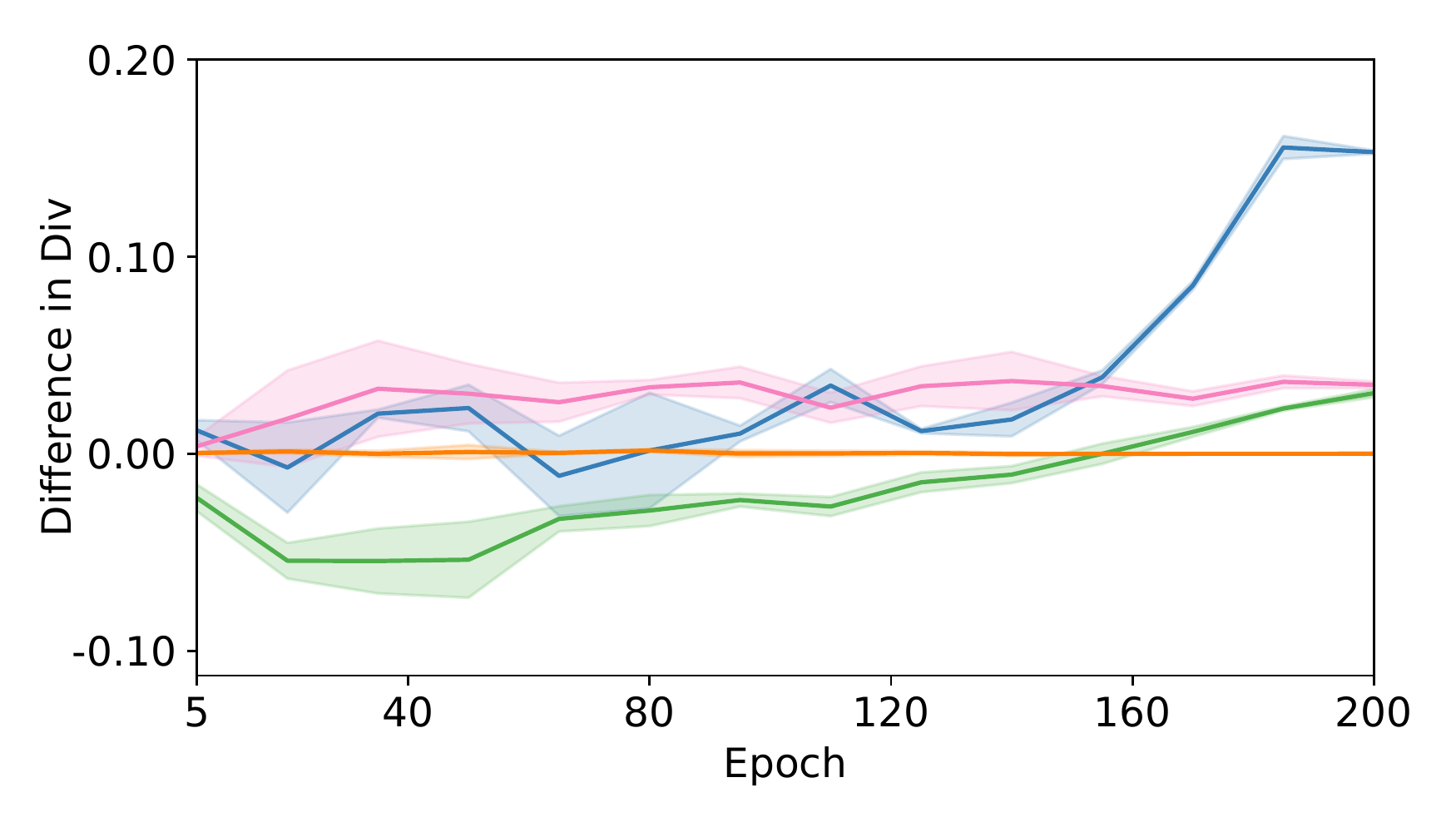}
    \caption{Diversification effects of various perturbations on the train data for the DE teachers (top) and the LatentBE students (bottom). The results are with WRN28x4 on CIFAR-100.}
    \label{fig:perturbation}
\end{figure}

\cref{fig:perturbation} shows the effects of the perturbation strategies during training. For this, we measure the changes in the function diversities of the teacher networks and student subnetworks due to a perturbation $\bvarepsilon$ during a training procedure:
\[
&\operatorname{TDiv}(\bsx + \bvarepsilon) - \operatorname{TDiv}(\bsx), \label{eq:tdiv_diff} \\
&\operatorname{SDiv}(\bsx + \bvarepsilon) - \operatorname{SDiv}(\bsx). \label{eq:sdiv_diff}
\]
Again, our intuition behind the perturbation strategy proposed in~\cref{main:sec:perturbation} is, pinpointing inputs that should be diversified for students while diversifying learning signals from teachers. In other words, the perturbation that decreases~\cref{eq:sdiv_diff} while increasing~\cref{eq:tdiv_diff} will be helpful for ensemble distillation.

\cref{fig:perturbation} clearly shows that our proposed perturbation (i.e., TDiv - Sdiv) accomplishes our goal to increase teacher diversities and decrease student diversities. Especially, the student diversities significantly drop during the early stage of training, but gradually increase as the training progress. We conjecture that this is because after enough training the students are diversified for a wide range of inputs, so the effect of $-\mathrm{Sdiv}$ perturbation gradually decreases. Although ConfODS exhibits high diversification effects on teachers as intended, it has no significant effect on the students.

As an ablation study, we compared the efficacy of the perturbation strategies in terms of the actual classification performance. \cref{table_ablation_perturb} shows that increasing student diversities (TDiv-SDiv) clearly improves the performance compared to the ones not considering the student diversities (ConfODS or TDiv).

\begin{table}
    \centering
    \caption{Ablation results for perturbation strategies. The results are with WRN28x4 on CIFAR-100.}
    \label{table_ablation_perturb}
    \vskip 0.1in
    \resizebox{\linewidth}{!}{
    \begin{tabular}{llllll}
    \toprule
    Method & ACC ($\uparrow$) & NLL ($\downarrow$) & ECE ($\downarrow$) & cNLL ($\downarrow$) & cECE ($\downarrow$) \\
    \midrule
    LatentBE-4  & 79.46$\spm{0.20}$ & 0.993$\spm{0.024}$ & 0.124$\spm{0.004}$ & 0.837$\spm{0.012}$ & 0.046$\spm{0.005}$ \\
    + ConfODS   & 79.27$\spm{0.30}$ & 0.955$\spm{0.008}$ & 0.115$\spm{0.002}$ & 0.840$\spm{0.007}$ & 0.048$\spm{0.004}$ \\
    + TDiv      & 79.40$\spm{0.03}$ & 0.861$\spm{0.000}$ & 0.084$\spm{0.002}$ & 0.819$\spm{0.001}$ & 0.046$\spm{0.001}$ \\
    + TDiv-SDiv & \textBF{80.02}$\spm{0.07}$ & \textBF{0.792}$\spm{0.004}$ & \textBF{0.067}$\spm{0.001}$ & \textBF{0.772}$\spm{0.003}$ & \textBF{0.041}$\spm{0.003}$ \\
    \bottomrule
    \end{tabular}}
\end{table}

\begin{table*}[t]
    \centering
    \caption{Results of the distilled students for WRN28x1 on CIFAR-10. Results with $\spm{\text{std.}}$ are averaged over 4 seeds.}
    \label{table_C10_LatentBE}
    \vskip 0.1in
    \setlength{\tabcolsep}{4pt}
    \resizebox{\linewidth}{!}{
    \begin{tabular}{lllllllllll}
    \toprule
    & \multicolumn{5}{c}{\textbf{\textit{Students distilled from DE-4 teacher}}} & \multicolumn{5}{c}{\textbf{\textit{Students distilled from DE-8 teacher}}} \\
    \cmidrule(lr){2-6}\cmidrule(lr){7-11}
    Method & ACC ($\uparrow$) & NLL ($\downarrow$) & ECE ($\downarrow$) & cNLL ($\downarrow$) & cECE ($\downarrow$) & ACC ($\uparrow$) & NLL ($\downarrow$) & ECE ($\downarrow$) & cNLL ($\downarrow$) & cECE ($\downarrow$) \\
    \midrule\midrule
    \multicolumn{11}{l}{\textbf{\textit{Single forward pass for inference:}}} \\
    \quad KD~\citep{hinton2015distilling}           & 93.70$\spm{0.10}$ & 0.270$\spm{0.004}$ & 0.042$\spm{0.001}$ & 0.201$\spm{0.003}$ & 0.011$\spm{0.002}$ & 93.68$\spm{0.15}$ & 0.273$\spm{0.007}$ & 0.042$\spm{0.002}$ & 0.202$\spm{0.003}$ & 0.010$\spm{0.001}$ \\
    \quad AE-KD~\citep{du2020agree}                 & 93.67$\spm{0.05}$ & 0.278$\spm{0.002}$ & 0.042$\spm{0.000}$ & 0.205$\spm{0.001}$ & 0.010$\spm{0.001}$ & 93.75$\spm{0.16}$ & 0.266$\spm{0.008}$ & 0.041$\spm{0.001}$ & 0.199$\spm{0.004}$ & 0.011$\spm{0.001}$ \\
    \quad Proxy-EnD$^2$~\citep{ryabinin2021scaling} & 93.67$\spm{0.04}$ & 0.270$\spm{0.005}$ & 0.042$\spm{0.001}$ & 0.200$\spm{0.002}$ & 0.011$\spm{0.001}$ & 94.04$\spm{0.12}$ & 0.263$\spm{0.003}$ & 0.039$\spm{0.001}$ & 0.195$\spm{0.002}$ & 0.009$\spm{0.001}$ \\
    \quad\textbf{KD + LatentBE (Ours)}             & \textBF{93.98}$\spm{0.20}$ & 0.263$\spm{0.003}$ & 0.041$\spm{0.002}$ & 0.194$\spm{0.002}$ & 0.011$\spm{0.002}$ & 93.97$\spm{0.10}$ & 0.263$\spm{0.004}$ & 0.041$\spm{0.002}$ & 0.194$\spm{0.002}$ & 0.012$\spm{0.001}$ \\
    \quad + ConfODS                                 & 93.95$\spm{0.12}$ & 0.223$\spm{0.007}$ & 0.032$\spm{0.001}$ & 0.186$\spm{0.004}$ & \textBF{0.008}$\spm{0.001}$ & 94.17$\spm{0.15}$ & 0.214$\spm{0.004}$ & 0.031$\spm{0.001}$ & 0.179$\spm{0.003}$ & 0.007$\spm{0.001}$ \\
    \quad + \textbf{TDiv-SDiv}                      & 93.95$\spm{0.01}$ & \textBF{0.205}$\spm{0.006}$ & \textBF{0.028}$\spm{0.001}$ & \textBF{0.181}$\spm{0.005}$ & \textBF{0.008}$\spm{0.002}$ & \textBF{94.19}$\spm{0.11}$ & \textBF{0.200}$\spm{0.004}$ & \textBF{0.027}$\spm{0.002}$ & \textBF{0.178}$\spm{0.002}$ & \textBF{0.006}$\spm{0.002}$ \\
    \midrule
    \multicolumn{11}{l}{\textbf{\textit{Multiple forward passes for inference:}}} \\
    \quad KD + BE~\citep{mariet2021distilling}      & 93.77$\spm{0.20}$ & 0.275$\spm{0.011}$ & 0.043$\spm{0.002}$ & 0.201$\spm{0.006}$ & 0.011$\spm{0.001}$ & 94.00$\spm{0.05}$ & 0.263$\spm{0.008}$ & 0.040$\spm{0.001}$ & 0.195$\spm{0.003}$ & 0.010$\spm{0.001}$ \\
    \quad + ConfODS~\citep{nam2021diversity}        & 94.06$\spm{0.13}$ & 0.223$\spm{0.007}$ & 0.031$\spm{0.002}$ & 0.188$\spm{0.004}$ & 0.006$\spm{0.002}$ & 94.06$\spm{0.10}$ & 0.222$\spm{0.003}$ & 0.032$\spm{0.001}$ & 0.185$\spm{0.002}$ & 0.008$\spm{0.001}$ \\
    \quad + TDiv-SDiv                               & 93.74$\spm{0.10}$ & 0.213$\spm{0.002}$ & 0.028$\spm{0.001}$ & 0.189$\spm{0.001}$ & 0.006$\spm{0.001}$ & 94.10$\spm{0.13}$ & 0.202$\spm{0.002}$ & 0.026$\spm{0.000}$ & 0.181$\spm{0.002}$ & 0.007$\spm{0.001}$ \\
    \bottomrule
    \end{tabular}}
    \vspace{1em}
    \caption{Results of the distilled students for WRN28x4 on CIFAR-100. Results with $\spm{\text{std.}}$ are averaged over 4 seeds.}
    \label{table_C100_LatentBE}
    \vskip 0.1in
    \setlength{\tabcolsep}{4pt}
    \resizebox{\linewidth}{!}{
    \begin{tabular}{lllllllllll}
    \toprule
    & \multicolumn{5}{c}{\textbf{\textit{Students distilled from DE-2 teacher}}} & \multicolumn{5}{c}{\textbf{\textit{Students distilled from DE-4 teacher}}} \\
    \cmidrule(lr){2-6}\cmidrule(lr){7-11}
    Method & ACC ($\uparrow$) & NLL ($\downarrow$) & ECE ($\downarrow$) & cNLL ($\downarrow$) & cECE ($\downarrow$) & ACC ($\uparrow$) & NLL ($\downarrow$) & ECE ($\downarrow$) & cNLL ($\downarrow$) & cECE ($\downarrow$) \\
    \midrule\midrule
    \multicolumn{11}{l}{\textbf{\textit{Single forward pass for inference:}}} \\
    \quad KD~\citep{hinton2015distilling}           & 79.15$\spm{0.25}$ & 1.029$\spm{0.008}$ & 0.125$\spm{0.002}$ & 0.864$\spm{0.006}$ & 0.046$\spm{0.004}$ & 79.09$\spm{0.20}$ & 1.038$\spm{0.011}$ & 0.130$\spm{0.003}$ & 0.861$\spm{0.007}$ & 0.046$\spm{0.001}$ \\
    \quad AE-KD~\citep{du2020agree}                 & 78.79$\spm{0.23}$ & 1.041$\spm{0.015}$ & 0.129$\spm{0.002}$ & 0.871$\spm{0.009}$ & 0.044$\spm{0.004}$ & 79.00$\spm{0.39}$ & 1.033$\spm{0.009}$ & 0.129$\spm{0.004}$ & 0.859$\spm{0.008}$ & 0.045$\spm{0.004}$ \\
    \quad Proxy-EnD$^2$~\citep{ryabinin2021scaling} & 78.40$\spm{0.28}$ & 1.072$\spm{0.012}$ & 0.138$\spm{0.003}$ & 0.894$\spm{0.007}$ & 0.047$\spm{0.002}$ & 78.75$\spm{0.28}$ & 1.076$\spm{0.016}$ & 0.138$\spm{0.002}$ & 0.886$\spm{0.011}$ & 0.046$\spm{0.003}$ \\
    \quad \textbf{KD + LatentBE (Ours)}             & 79.16$\spm{0.18}$ & 0.985$\spm{0.011}$ & 0.122$\spm{0.002}$ & 0.848$\spm{0.006}$ & 0.045$\spm{0.003}$ & 79.46$\spm{0.20}$ & 0.993$\spm{0.024}$ & 0.124$\spm{0.004}$ & 0.837$\spm{0.012}$ & 0.046$\spm{0.005}$ \\
    \quad + ConfODS                                 & 78.61$\spm{0.19}$ & 0.965$\spm{0.008}$ & 0.109$\spm{0.001}$ & 0.873$\spm{0.006}$ & 0.046$\spm{0.002}$ & 79.27$\spm{0.30}$ & 0.955$\spm{0.008}$ & 0.115$\spm{0.002}$ & 0.840$\spm{0.007}$ & 0.048$\spm{0.004}$ \\
    \quad + \textbf{TDiv-SDiv}                      & \textBF{79.49}$\spm{0.15}$ & \textBF{0.826}$\spm{0.007}$ & \textBF{0.072}$\spm{0.003}$ & \textBF{0.798}$\spm{0.005}$ & \textBF{0.041}$\spm{0.002}$ & \textBF{80.02}$\spm{0.07}$ & \textBF{0.792}$\spm{0.004}$ & \textBF{0.067}$\spm{0.001}$ & \textBF{0.772}$\spm{0.003}$ & \textBF{0.041}$\spm{0.003}$ \\
    \midrule
    \multicolumn{11}{l}{\textbf{\textit{Multiple forward passes for inference:}}} \\
    \quad KD + BE~\citep{mariet2021distilling}      & 78.50$\spm{0.42}$ & 1.067$\spm{0.010}$ & 0.134$\spm{0.003}$ & 0.888$\spm{0.008}$ & 0.044$\spm{0.003}$ & 78.92$\spm{0.24}$ & 1.035$\spm{0.013}$ & 0.130$\spm{0.002}$ & 0.863$\spm{0.012}$ & 0.043$\spm{0.002}$ \\
    \quad + ConfODS~\citep{nam2021diversity}        & 78.24$\spm{0.19}$ & 1.011$\spm{0.013}$ & 0.118$\spm{0.002}$ & 0.897$\spm{0.005}$ & 0.048$\spm{0.003}$ & 78.65$\spm{0.29}$ & 1.002$\spm{0.013}$ & 0.123$\spm{0.002}$ & 0.873$\spm{0.010}$ & 0.046$\spm{0.003}$ \\
    \quad + TDiv-SDiv                               & 78.50$\spm{0.21}$ & 0.871$\spm{0.005}$ & 0.080$\spm{0.003}$ & 0.837$\spm{0.004}$ & 0.044$\spm{0.002}$ & 79.56$\spm{0.18}$ & 0.818$\spm{0.007}$ & 0.066$\spm{0.002}$ & 0.798$\spm{0.006}$ & 0.042$\spm{0.003}$ \\
    \bottomrule
    \end{tabular}}
\end{table*}

\subsection{Results on CIFAR-10/100}\label{main:subsec:classification}

We compare ours to the existing ensemble distillation methods. Here, we consider baselines using a single student network: KD~\citep{hinton2015distilling}, AE-KD~\citep{du2020agree}, and Proxy-EnD$^2$~\citep{ryabinin2021scaling}. As suggested in \citet{ashukha2020pitfalls}, we report both original and calibrated metrics for NLL and ECE. See~\cref{app:subsec:distillation_methods} for the details in distillation methods, and \cref{app:evaluation} for the definitions of evaluation metrics.

The results for CIFAR-10 and CIFAR-100 are presented in~\cref{table_C10_LatentBE,table_C100_LatentBE}. LatentBE consistently outperforms the baselines both in terms of accuracy and uncertainty estimates, and our perturbation further boosts up the performance of LatentBE. We note that our method gets better as the number of teachers increases, indicating that ours effectively transfers diversities from multiple teachers.

Moreover, a benefit of our approach is that it consistently outperforms the vanilla KD even when the number of teachers $M$ is small. On the other hand, \citet{ryabinin2021scaling} suffers when $M$ is small because the estimation for Dirichlet parameters required for the distillation become inaccurate. This gives an advantage to our method under resource limited setting where training large number of ensemble teachers are intractable.

\begin{table}[t]
    \caption{Results of the distilled students for WRN28x1 on CIFAR-10-C, and WRN28x4 on CIFAR-100-C. The results are with the DE-4 teacher and are averaged over 4 seeds.}
    \label{table_CIFAR_Corrupted}
    \vskip 0.1in
    \centering
    \setlength{\tabcolsep}{4pt}
    \resizebox{\linewidth}{!}{
    \begin{tabular}{llll}
    \toprule
    Method & ACC ($\uparrow$) & NLL ($\downarrow$) & ECE ($\downarrow$) \\
    \midrule\midrule
    \multicolumn{4}{l}{\textbf{CIFAR-10-C:}} \\
    \quad \underline{DE-4 teacher}                  & \underline{73.18} & \underline{1.025}  & \underline{0.092}  \\
    \quad KD~\citep{hinton2015distilling}           & 72.49$\spm{0.38}$ & 1.492$\spm{0.039}$ & 0.202$\spm{0.004}$ \\
    \quad AE-KD~\citep{du2020agree}                 & 72.06$\spm{0.49}$ & 1.540$\spm{0.046}$ & 0.207$\spm{0.005}$ \\
    \quad Proxy-EnD$^2$~\citep{ryabinin2021scaling} & 71.05$\spm{0.51}$ & 1.600$\spm{0.050}$ & 0.217$\spm{0.007}$ \\
    \quad \textbf{KD + LatentBE (Ours)}             & 72.73$\spm{0.54}$ & 1.522$\spm{0.078}$ & 0.203$\spm{0.008}$ \\
    \quad + ConfODS                                 & 70.18$\spm{0.48}$ & 1.477$\spm{0.038}$ & 0.200$\spm{0.004}$ \\
    \quad + \textbf{TDiv-SDiv}                      & \textBF{73.22}$\spm{0.43}$ & \textBF{1.237}$\spm{0.024}$ & \textBF{0.171}$\spm{0.004}$ \\
    \midrule
    \multicolumn{4}{l}{\textbf{CIFAR-100-C:}} \\
    \quad \underline{DE-4 teacher}                  & \underline{51.08} & \underline{2.296}  & \underline{0.114}  \\
    \quad KD~\citep{hinton2015distilling}           & 48.79$\spm{0.16}$ & 3.410$\spm{0.056}$ & 0.341$\spm{0.005}$ \\
    \quad AE-KD~\citep{du2020agree}                 & 48.86$\spm{0.12}$ & 3.407$\spm{0.066}$ & 0.340$\spm{0.007}$ \\
    \quad Proxy-EnD$^2$~\citep{ryabinin2021scaling} & 48.67$\spm{0.25}$ & 3.378$\spm{0.054}$ & 0.350$\spm{0.004}$ \\
    \quad \textbf{KD + LatentBE (Ours)}             & 50.03$\spm{0.36}$ & 3.178$\spm{0.053}$ & 0.321$\spm{0.003}$ \\
    \quad + ConfODS                                 & 47.17$\spm{0.12}$ & 3.273$\spm{0.019}$ & 0.326$\spm{0.002}$ \\
    \quad + \textbf{TDiv-SDiv}                      & \textBF{51.36}$\spm{0.28}$ & \textBF{2.507}$\spm{0.059}$ & \textBF{0.208}$\spm{0.006}$ \\
    \bottomrule
    \end{tabular}}
\end{table}

\paragraph{Robustness to common corruptions}
We further evaluate our method on corrupted CIFAR datasets to verify its robustness under common corruptions~\citep{hendrycks2019benchmarking}. \cref{table_CIFAR_Corrupted} reports the evaluation metrics averaged over all types of corruptions and intensities and shows that ours are better calibrated than the existing baselines.

\begin{figure*}[t]
    \centering
    \subfigure{\includegraphics[width=\linewidth]{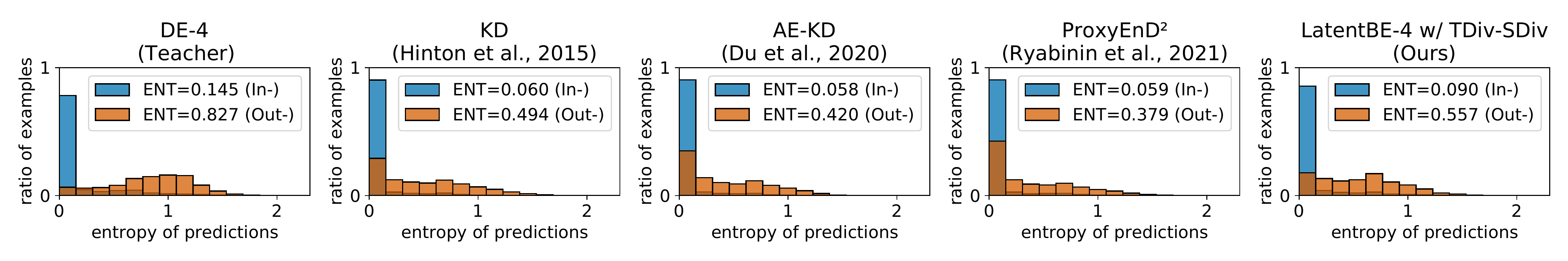}}\vspace*{-1cm}
    \caption{Histograms of the predictive entropy for WRN28x1 students distilled from DE-4 on CIFAR-10. `In-' denotes the entropy on the in-distribution data (i.e., CIFAR-10), and `Out-' denotes the entropy on the out-of-distribution data (i.e., SVHN).}
    \label{fig:ood_c10}
    \centering
    \subfigure{\includegraphics[width=\linewidth]{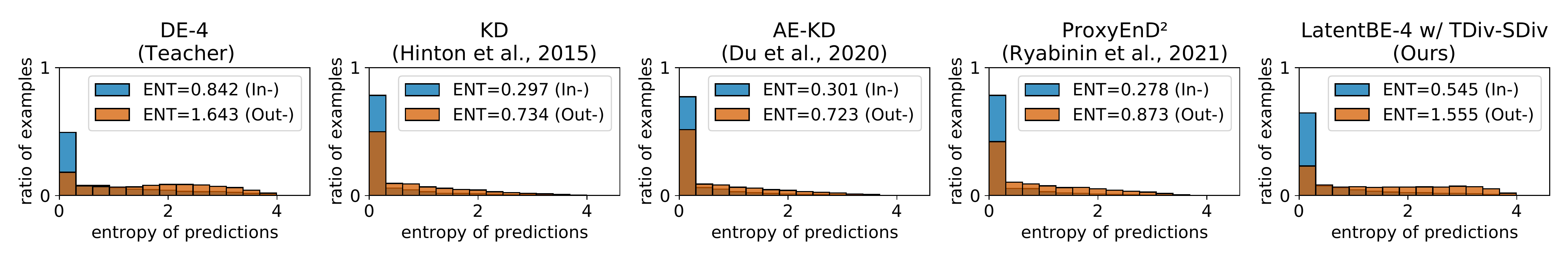}}\vspace*{-1cm}
    \caption{Histograms of the predictive entropy for WRN28x4 students distilled from DE-4 on CIFAR-100. `In-' denotes the entropy on the in-distribution data (i.e., CIFAR-100), and `Out-' denotes the entropy on the out-of-distribution data (i.e., SVHN).}
    \label{fig:ood_c100}
\end{figure*}

\paragraph{Out-of-distribution data}
A well-calibrated classifier should be uncertain for out-of-distribution data while being certain for in-distribution data. Following~\citet{lakshminarayanan2017simple}, we draw histograms depicting the distribution of the predictive entropy for in-distribution (i.e., CIFAR-10/100) and out-of-distribution data (i.e., SVHN) in~\cref{fig:ood_c10,fig:ood_c100}. It shows that ours are more uncertain about unseen classes, behaving similarly to ensemble teachers. Refer to~\cref{app:subsec:uncertainty} for more results with other out-of-distribution datasets.

\subsection{Runtime analysis}
\cref{table_C10_LatentBE,table_C100_LatentBE} show that LatentBE is competitive or even better for some metrics than the one-to-one distillation methods with BE students~\citep{mariet2021distilling, nam2021diversity}. This is quite remarkable since BE requires additional inference cost while LatentBE doesn't. More specifically, \cref{table_costs} reports the runtimes of BE and LatentBE on the same single GeForce RTX 3090 setting. We measure the wall-clock time in hours for training and milliseconds per image (ms/img) for inference. In principle, BE requires multiple forward passes for inference (i.e., 0.390 ms/img), but this can be avoided by parallelized inference (0.288 ms/img). Still, the parallelization requires additional cost from handling larger mini-batches, which is still significantly larger than a single model inference time (0.107 ms/img).

\begin{table}[t]
    \centering
    \caption{Comparison on training and testing runtime. The results are with WRN28x4 on CIFAR-100 previously reported in~\cref{table_C100_LatentBE}.}
    \label{table_costs}
    \vskip 0.1in
    \resizebox{\linewidth}{!}{\begin{tabular}{lrrrr}
    \toprule
    & \multicolumn{2}{c}{\textbf{\textit{Performance}}} & \multicolumn{2}{c}{\textbf{\textit{Runtime}}} \\
    \cmidrule(lr){2-3}\cmidrule(lr){4-5}
    Method & ACC ($\uparrow$) & NLL ($\downarrow$) & Training ($\downarrow$) & Inference ($\downarrow$) \\
    \midrule
    DE-4 teacher                       & 81.37 & 0.706 & 6.2 hrs. & 0.390 ms/img \\
    \midrule
    KD~\citep{hinton2015distilling}    & 79.09 & 1.038 & 1.6 hrs. & 0.107 ms/img \\
    \textbf{KD + LatentBE (Ours)}      & 79.46 & 0.993 & 3.3 hrs. & 0.107 ms/img \\
    \textbf{+ TDiv-SDiv}               & 80.02 & 0.792 & 5.3 hrs. & 0.107 ms/img \\
    \midrule
    KD + BE~\citep{mariet2021distilling}     & 78.92 & 1.035 & 3.3 hrs. & 0.288 ms/img \\
    + TDiv-SDiv                        & 79.56 & 0.818 & 5.3 hrs. & 0.288 ms/img \\
    \bottomrule
    \end{tabular}}
\end{table}

\subsection{Results on TinyImageNet and ImageNet-1k}
We verify the scalability of our proposed method on large-scale datasets, including TinyImageNet and ImageNet-1k~\citep{russakovsky2015imagenet}. \cref{table_TIN200_LatentBE,table_ImageNet1k_LatentBE} show that ours outperform the baselines for both datasets.
\begin{table*}[t]
    \centering
    \caption{Results of R18 and WRN28x4 on TinyImageNet. Results with $\spm{\text{std.}}$ are averaged over 3 seeds.}
    \label{table_TIN200_LatentBE}
    \vskip 0.1in
    \setlength{\tabcolsep}{4pt}
    \resizebox{\linewidth}{!}{
    \begin{tabular}{lllllllllll}
    \toprule
    & \multicolumn{5}{c}{\textbf{\textit{R18 on TinyImageNet}}} & \multicolumn{5}{c}{\textbf{\textit{WRN28x4 on TinyImageNet}}} \\
    \cmidrule(lr){2-6}\cmidrule(lr){7-11}
    Method & ACC ($\uparrow$) & NLL ($\downarrow$) & ECE ($\downarrow$) & cNLL ($\downarrow$) & cECE ($\downarrow$) & ACC ($\uparrow$) & NLL ($\downarrow$) & ECE ($\downarrow$) & cNLL ($\downarrow$) & cECE ($\downarrow$) \\
    \midrule\midrule
    \multicolumn{11}{l}{\textbf{\textit{Baseline results:}}} \\
    \quad Base (w/o distillation)                & 64.78$\spm{0.12}$ & 1.580$\spm{0.004}$ & 0.104$\spm{0.002}$ & 1.494$\spm{0.001}$ & 0.031$\spm{0.002}$ & 63.25$\spm{0.07}$ & 1.599$\spm{0.010}$ & 0.099$\spm{0.003}$ & 1.517$\spm{0.006}$ & 0.022$\spm{0.001}$ \\
    \quad Single KD~\citep{hinton2015distilling} & 67.29$\spm{0.16}$ & 1.470$\spm{0.005}$ & 0.107$\spm{0.002}$ & 1.383$\spm{0.003}$ & 0.046$\spm{0.002}$ & 66.25$\spm{0.38}$ & 1.443$\spm{0.012}$ & 0.077$\spm{0.005}$ & 1.391$\spm{0.007}$ & 0.027$\spm{0.002}$ \\
    \quad DE-4 teacher                           & 69.28 & 1.273 & 0.025 & 1.272 & 0.023 & 68.75 & 1.276 & 0.027 & 1.277 & 0.027 \\
    \midrule
    \multicolumn{11}{l}{\textbf{\textit{Students distilled from DE-4 teacher:}}} \\
    \quad KD~\citep{hinton2015distilling}           & 68.88$\spm{0.20}$ & 1.391$\spm{0.009}$ & 0.099$\spm{0.001}$ & 1.317$\spm{0.006}$ & 0.044$\spm{0.002}$ & 67.69$\spm{0.08}$ & 1.351$\spm{0.004}$ & 0.067$\spm{0.002}$ & 1.314$\spm{0.004}$ & 0.026$\spm{0.006}$ \\
    \quad AE-KD~\citep{du2020agree}                 & 67.20$\spm{0.08}$ & 1.409$\spm{0.006}$ & 0.089$\spm{0.001}$ & 1.355$\spm{0.005}$ & 0.040$\spm{0.003}$ & 64.77$\spm{0.08}$ & 1.424$\spm{0.004}$ & 0.052$\spm{0.002}$ & 1.403$\spm{0.004}$ & 0.020$\spm{0.006}$ \\
    \quad Proxy-EnD$^2$~\citep{ryabinin2021scaling} & 62.42$\spm{0.26}$ & 1.572$\spm{0.002}$ & \textBF{0.017}$\spm{0.003}$ & 1.571$\spm{0.002}$ & \textBF{0.021}$\spm{0.004}$ & 62.29$\spm{0.25}$ & 1.578$\spm{0.003}$ & \textBF{0.049}$\spm{0.004}$ & 1.560$\spm{0.005}$ & \textBF{0.016}$\spm{0.003}$ \\
    \quad \textbf{KD + LatentBE (Ours)}             & 68.96$\spm{0.19}$ & 1.391$\spm{0.007}$ & 0.103$\spm{0.002}$ & 1.312$\spm{0.007}$ & 0.042$\spm{0.002}$ & 67.76$\spm{0.05}$ & 1.343$\spm{0.004}$ & 0.073$\spm{0.001}$ & 1.303$\spm{0.004}$ & 0.025$\spm{0.001}$ \\
    \quad + ConfODS                                 & 69.00$\spm{0.32}$ & 1.390$\spm{0.008}$ & 0.101$\spm{0.004}$ & 1.313$\spm{0.005}$ & 0.047$\spm{0.005}$ & 67.89$\spm{0.18}$ & 1.338$\spm{0.005}$ & 0.071$\spm{0.003}$ & 1.299$\spm{0.004}$ & 0.027$\spm{0.002}$ \\
    \quad + \textbf{TDiv-SDiv}                      & \textBF{69.14}$\spm{0.27}$ & \textBF{1.342}$\spm{0.005}$ & 0.085$\spm{0.004}$ & \textBF{1.290}$\spm{0.001}$ & 0.038$\spm{0.002}$ & \textBF{68.15}$\spm{0.03}$ & \textBF{1.317}$\spm{0.008}$ & 0.062$\spm{0.000}$ & \textBF{1.286}$\spm{0.006}$ & 0.027$\spm{0.002}$ \\
    \bottomrule
    \end{tabular}}
\end{table*}
\begin{table}[t]
    \centering
    \caption{Results for R50 on ImageNet-1k.}
    \label{table_ImageNet1k_LatentBE}
    \vskip 0.1in
    \setlength{\tabcolsep}{4pt}
    \resizebox{\linewidth}{!}{\begin{tabular}{llllll}
    \toprule
    & \multicolumn{5}{c}{\textbf{\textit{R50 on ImageNet-1k}}} \\
    \cmidrule(lr){2-6}
    Method & ACC ($\uparrow$) & NLL ($\downarrow$) & ECE ($\downarrow$) & cNLL ($\downarrow$) & cECE ($\downarrow$) \\
    \midrule\midrule
    \multicolumn{6}{l}{\textbf{\textit{Baselines results:}}} \\
    \quad Base (w/o distillation)                & 76.80 & 0.927 & 0.040 & 0.913 & 0.019 \\
    \quad Single KD~\citep{hinton2015distilling} & 76.90 & 0.918 & 0.028 & 0.913 & 0.017 \\
    \quad DE-2 teacher                           & 77.96 & 0.862 & 0.018 & 0.859 & 0.018 \\
    \midrule
    \multicolumn{6}{l}{\textbf{\textit{Students distilled from DE-2 teacher:}}} \\
    \quad KD~\citep{hinton2015distilling}        & 77.01 & 0.904 & 0.028 & 0.900 & \textBF{0.017} \\
    \quad \textbf{KD + LatentBE (Ours)}          & 77.30 & 0.902 & 0.029 & 0.895 & 0.019 \\
    \quad \textbf{+ TDiv-SDiv}                   & \textBF{77.38} & \textBF{0.898} & \textBF{0.027} & \textBF{0.892} & 0.018 \\
    \bottomrule
    \end{tabular}}
\end{table}

\section{Conclusion}
In this paper, we proposed a novel ensemble distillation algorithm improving both prediction accuracy and uncertainty calibration without increasing inference cost. We first presented LatentBE, where the rank-one factors of BEs are trained in a one-to-one way, but later weight-averaged for inference. We showed that under a suitable training scheme, the subspaces defined by the rank-one factors of BE remain in a flat minimum, and weight averaging those rank-one factors thus yields a robust single student model. We further presented a novel perturbation strategy for ensemble distillation that decreases student diversities and increases teacher diversities at the same time. By training with inputs perturbed in that way, we can effectively enhance the diversities of students. Our ensemble distillation algorithm combining these two achieved remarkable performance on various image classification tasks. 

\section*{Acknowledgements}
This work was partly supported by KAIST-NAVER Hypercreative AI Center, Institute of Information \& communications Technology Planning \& Evaluation (IITP) grant funded by the Korea government (MSIT) (No.2019-0-00075, Artificial Intelligence Graduate School Program (KAIST), No. 2021-0-02068, Artificial Intelligence Innovation Hub), and National Research Foundation of Korea (NRF) funded by the Ministry of Education (NRF-2021M3E5D9025030).
\bibliography{references}
\bibliographystyle{icml2022}

\newpage
\appendix
\onecolumn

\section{Additional Results}
\subsection{Baseline results for CIFAR-10/100}
In~\cref{table_CIFAR_DE}, we report the baseline results for the experiments on CIFAR-10/100: (1) evaluation results for DE teachers distilling knowledge into students, (2) performance of the single model trained with the classical cross-entropy loss without distillation, and (3) performance of the single model distilled from DE-1.
\begin{table*}[t]
    \caption{Baseline results for the experiments on CIFAR-10/100. Results with $\spm{\text{std.}}$ are averaged over 4 seeds.}
    \label{table_CIFAR_DE}
    \vskip 0.1in
    \centering
    \setlength{\tabcolsep}{4pt}
    \resizebox{\linewidth}{!}{
    \begin{tabular}{lllllllllll}
    \toprule
    & \multicolumn{5}{c}{\textbf{WRN28x1 on CIFAR-10}} & \multicolumn{5}{c}{\textbf{WRN28x4 on CIFAR-100}} \\
    \cmidrule(lr){2-6}\cmidrule(lr){7-11}
    Method & ACC ($\uparrow$) & NLL ($\downarrow$) & ECE ($\downarrow$) & cNLL ($\downarrow$) & cECE ($\downarrow$) & ACC ($\uparrow$) & NLL ($\downarrow$) & ECE ($\downarrow$) & cNLL ($\downarrow$) & cECE ($\downarrow$) \\
    \midrule\midrule
    DE-1 & 93.01 & 0.271 & 0.038 & 0.222 & 0.009 & 78.03 & 0.888 & 0.062 & 0.878 & 0.041 \\
    DE-2 & 94.11 & 0.203 & 0.016 & 0.191 & 0.010 & 80.13 & 0.768 & 0.026 & 0.768 & 0.025 \\
    DE-3 & 94.53 & 0.180 & 0.011 & 0.176 & 0.012 & 80.93 & 0.723 & 0.026 & 0.720 & 0.022 \\
    DE-4 & 94.71 & 0.170 & 0.006 & 0.168 & 0.010 & 81.37 & 0.706 & 0.031 & 0.700 & 0.018 \\
    DE-5 & 94.68 & 0.166 & 0.008 & 0.165 & 0.009 & -     & -     & -     & -     & -     \\
    DE-6 & 94.64 & 0.162 & 0.007 & 0.162 & 0.007 & -     & -     & -     & -     & -     \\
    DE-7 & 94.89 & 0.159 & 0.008 & 0.159 & 0.008 & -     & -     & -     & -     & -     \\
    DE-8 & 95.07 & 0.157 & 0.009 & 0.157 & 0.009 & -     & -     & -     & -     & -     \\
    \midrule
    Base (w/o distillation)                & 93.05$\spm{0.16}$ & 0.263$\spm{0.008}$ & 0.037$\spm{0.002}$ & 0.218$\spm{0.005}$ & 0.006$\spm{0.002}$ & 77.93$\spm{0.21}$ & 0.893$\spm{0.006}$ & 0.060$\spm{0.002}$ & 0.882$\spm{0.006}$ & 0.041$\spm{0.003}$ \\
    Single KD~\citep{hinton2015distilling} & 93.52$\spm{0.09}$ & 0.274$\spm{0.008}$ & 0.041$\spm{0.001}$ & 0.206$\spm{0.004}$ & 0.009$\spm{0.001}$ & 78.47$\spm{0.13}$ & 1.039$\spm{0.005}$ & 0.127$\spm{0.003}$ & 0.886$\spm{0.003}$ & 0.045$\spm{0.002}$ \\
    \bottomrule
    \end{tabular}}
\end{table*}

\subsection{Further experiments on predictive uncertainty}\label{app:subsec:uncertainty}
\begin{figure*}[t]
    \centering
    \subfigure{\includegraphics[width=\linewidth]{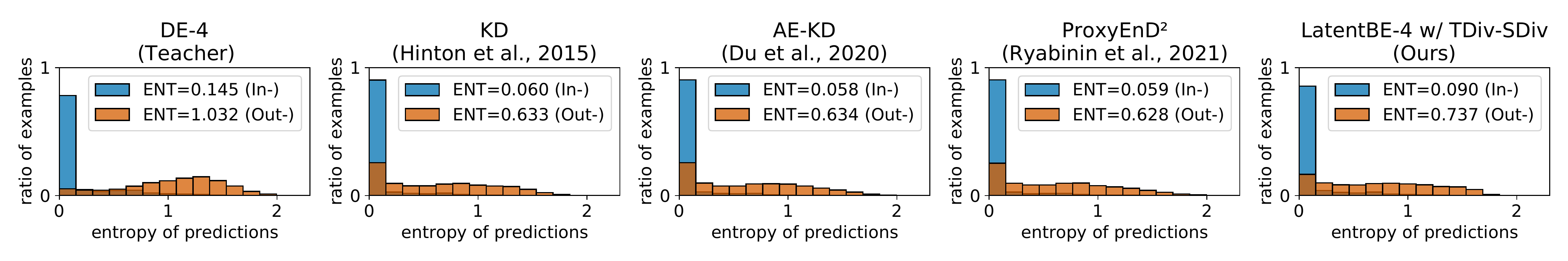}}\vspace*{-0.5cm}
    \subfigure{\includegraphics[width=\linewidth]{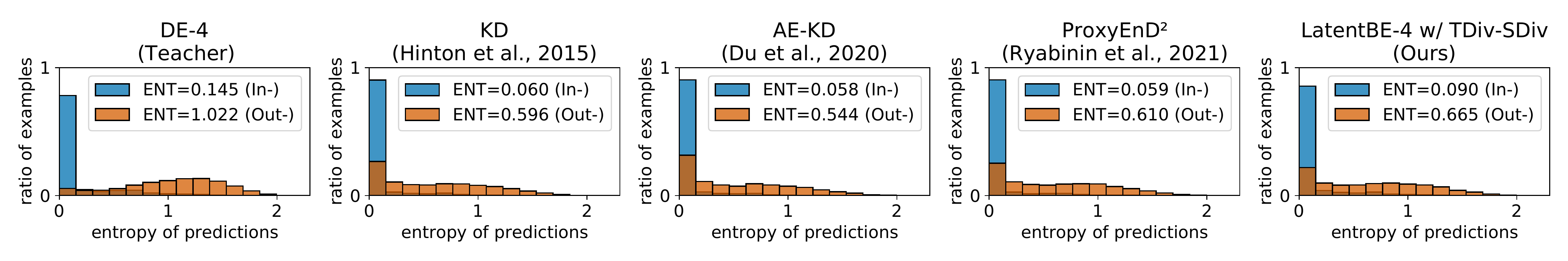}}\vspace*{-0.5cm}
    \caption{Histograms of the predictive entropy for WRN28x1 students distilled from DE-4 on CIFAR-10. `In-' denotes the entropy on the in-distribution data (i.e., CIFAR-10), and `Out-' denotes the entropy on the out-of-distribution data including LSUN (1st row) and TinyImageNet (2nd row).}
    \label{fig:ood_c10_full}
    \centering
    \subfigure{\includegraphics[width=\linewidth]{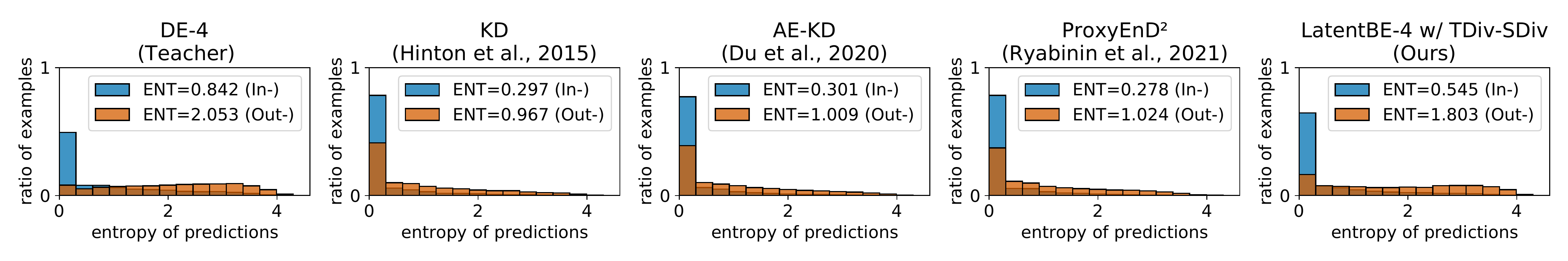}}\vspace*{-0.5cm}
    \subfigure{\includegraphics[width=\linewidth]{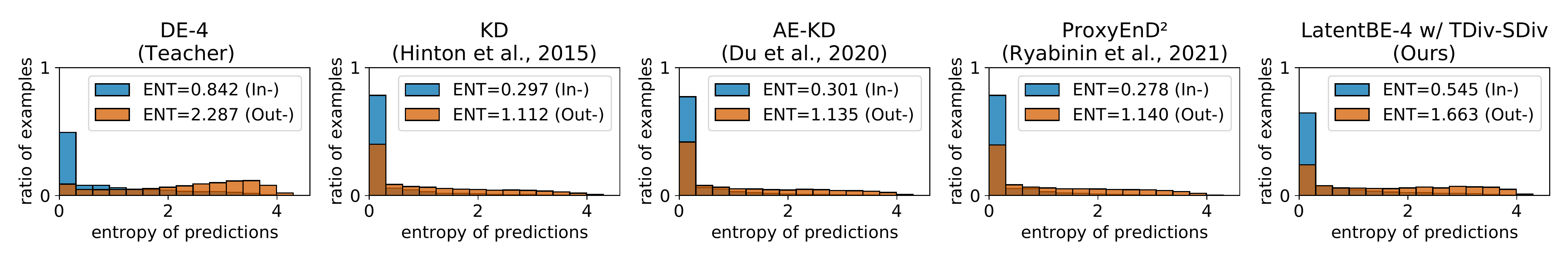}}\vspace*{-0.5cm}
    \caption{Histograms of the predictive entropy for WRN28x4 students distilled from DE-4 on CIFAR-100. `In-' denotes the entropy on the in-distribution data (i.e., CIFAR-100), and `Out-' denotes the entropy on the out-of-distribution data including LSUN (1st row) and TinyImageNet (2nd row).}
    \label{fig:ood_c100_full}
\end{figure*}

For the experiments on predictive uncertainty, we consider SVHN~\citep{netzer2011reading}, LSUN~\citep{yu2015lsun}, and TinyImageNet~\citep{russakovsky2015imagenet} as out-of-distribution data. Here, LSUN and TinyImageNet images are downscaled into $32\times32\times3$. \cref{fig:ood_c10_full,fig:ood_c100_full} further provide the predictive uncertainty results on LSUN and TinyImageNet. Again, our approach exhibits higher predictive uncertainty on out-of-distribution examples than existing baselines.

\section{Experimental Details}\label{app:experimental_details}
Code is available at \href{https://github.com/cs-giung/distill-latentbe}{https://github.com/cs-giung/distill-latentbe}. Our implementation for the experiments on CIFAR-10/100 and TinyImageNet are built on PyTorch~\citep{paszke2019pytorch}. Besides, the experiments on ImageNet-1k are conducted with 8 TPUv3 cores, supported by the TPU Research Cloud\footnote{\href{https://sites.research.google/trc/about/}{https://sites.research.google/trc/about/}}.

\subsection{Datasets}

\paragraph{CIFAR-10/100}
The dataset is available at~\href{https://www.cs.toronto.edu/~kriz/cifar.html}{https://www.cs.toronto.edu/ \textasciitilde kriz/cifar.html}. It consists of 50,000 train examples and 10,000 test examples from 10/100 classes, with images size of $32\times32\times3$. In this paper, the last 5,000 examples of the train split are used as the validation split for computing calibrated metrics. We follow the standard data augmentation policy~\citep{he2016deep} which consists of random cropping of 32 pixels with a padding of 4 pixels and random horizontal flipping. Throughout experiments on CIFAR-10/100 classification, we use WideResNet (WRN) networks introduced in~\citet{zagoruyko2016wrn}; WRN28x1 on CIFAR-10 and WRN28x4 on CIFAR-100.

\paragraph{TinyImageNet}
The dataset is available at~\href{http://cs231n.stanford.edu/tiny-imagenet-200.zip}{http://cs231n.stanford.edu/tiny-imagenet-200.zip}. It consists of 100,000 train examples, 10,000 validation examples and 10,000 test examples from 200 classes subsampled from ImageNet-1k, with images size of $64\times64\times3$. Since the labels of the official test set are not publicly available, we use the official validation set as a test set for experiments. Consequently, the last 500 examples for each class of the train split are used as the validation split for computing calibrated metrics, i.e., train and validation split consists of 90,000 and 10,000 examples, respectively. We apply the data augmentation which consists of random cropping of 64 pixels with a padding of 4 pixels and random horizontal flipping. Throughout experiments on TinyImageNet classification, we use the ResNet-18 (R18) network introduced in~\citet{he2016deep} and the WRN28x4 network introduced in~\citet{zagoruyko2016wrn}.

\paragraph{ImageNet-1k}
It consists of 1,281,167 train examples, 50,000 validation examples and 100,000 test images from 1,000 classes. Since the labels of the official test set are not publicly available, we only report the evaluation results on the validation set. We follow the standard data augmentation policy from PyTorch which consists of random cropping with an images size of $224\times224\times3$ and random horizontal flipping\footnote{\href{https://github.com/pytorch/examples/tree/master/imagenet}{https://github.com/pytorch/examples/tree/master/imagenet}}. Throughout experiments on ImageNet-1k classification, we use the ResNet-50 (R50) network introduced in~\citet{he2016deep}.

\subsection{Training details}
All images are standardized by subtracting the per-channel mean and dividing the result by the per-channel standard deviation. We use SGD optimizer with Nesterov momentum $0.9$, and a single-cycle cosine annealing learning rate schedule with a linear warm-up, i.e., the learning rate starts from $0.01\times\texttt{base\_lr}$ and reaches $\texttt{base\_lr}$ after the first $5$ epochs, and is decayed by the single-cycle cosine annealing learning rate schedule. More precisely, (1) for CIFAR-10/100, we run $200$ epochs on a single machine with batch size $128$ and $\texttt{base\_lr}=0.1$, (2) for TinyImageNet, we run $80$ epochs on four machines with the total batch size $128$ and $\texttt{base\_lr}=0.1$, and (3) for ImageNet-1k, we run $100$ epochs on eight machines with the total batch size $256$ and $\texttt{base\_lr}=0.1$. We also apply the weight decay~\citep{krogh1991simple} to regularize training; the weight decay coefficient is set to be $0.0005$ for CIFAR-10/100 and TinyImageNet, and $0.0001$ for ImageNet-1k.

\subsection{Evaluation}
\label{app:evaluation}
The problem we address in this paper is the $K$-way classification problem; a neural network $\bsf : \bbR^D \rightarrow \bbR^K$ takes $D$-dimensional inputs $\bsx$ (i.e., images) and makes predictions about outputs $y$ (i.e., class label) with $K$-dimensional logits.
We denote the output probabilities of the model $\bsf$ for a given input $\bsx$ as
\[
\bsp_{\bsf}^{(k)}(\bsx) = \frac{
    \exp{\left( \bsf^{(k)}(\bsx) \right)}
}{
    \sum_{j=1}^{K} \exp{\left( \bsf^{(j)}(\bsx) \right)}
}, \quad \text{for } k=1,...,K.
\]

\paragraph{Standard metrics}
We evaluate the following \textit{standard metrics} of the model $\bsf$ on the dataset $\calD$:
\begin{itemize}
\item ACC (accuracy; higher is better):
\[
\operatorname{ACC}(\bsf, \calD) =
\frac{1}{|\calD|} \sum_{(\bsx,y)\in\calD} \left[
    y = \argmax_k \bsp_{\bsf}^{(k)}(\bsx)
\right],
\]
where $\left[ \cdot \right]$ denotes the Iverson bracket.
\item NLL (negative log-likelihood; lower is better):
\[
\operatorname{NLL}(\bsf, \calD) =
- \sum_{(\bsx,y)\in\calD} y \log{\bsp_{\bsf}^{(y)}(\bsx)}
\]
\item ECE (expected calibration error; lower is better):
\[
\operatorname{ECE}(\bsf, \calD, L) = \sum_{l=1}^{L} \frac{|\calB_l|}{|\calD|} \left\lvert
    \operatorname{ACC(\bsf, \calB_l)} - \sum_{(\bsx,\cdot)\in\calB_l} \frac{\max_{k}\bsp_{\bsf}^{(k)}(\bsx)}{|\calB_l|}
\right\rvert,
\]
where $\{\calB_1,...,\calB_L\}$ is a partition of $\calD$, where $\calB_l = \{(\bsx,y)\in\calD \;|\; \max_{k} \bsp_{\bsf}^{(k)}(\bsx) \in ( (l-1)/L, l/L ] \}$.
Here, the difference between the accuracy and mean confidence of predictions represents the calibration gap for each bin.
We fixed $L=15$ for all evaluation results in this paper.
\end{itemize}

\paragraph{Calibrated metrics}
Temperature scaling softens output probabilities of the model with a single scale parameter $\tau>0$, and it can be used for calibrating probabilistic models \textit{without affecting the model's prediction accuracy}~\citep{guo2017calibration}.
We define the temperature scaled output probabilities of the model $\bsf$ for a given input $\bsx$ as
\[
\bsp_{\bsf}^{(k)}(\bsx;\tau) = \frac{
    \exp{\left( \bsf^{(k)}(\bsx) / \tau \right)}
}{
    \sum_{j=1}^{K} \exp{\left( \bsf^{(j)}(\bsx) / \tau \right)}
}, \quad \text{for } k=1,...,K.
\]
Following \citet{ashukha2020pitfalls}, we also evaluate the \textit{calibrated metrics} which are computed using the temperature scaled outputs.
Specifically, we first find the optimal temperature which minimizes the NLL on the validation split $\calD_{\text{valid}}$, i.e.,
\[
\tau^\ast \gets \argmin_\tau \left[
    - \sum_{(\bsx,y)\in\calD_{\text{valid}}} y \log{\bsp_{\bsf}^{(y)}(\bsx;\tau)}
\right],
\]
and then we can compute the following \textit{calibrated metrics}:
\begin{itemize}
\item cNLL (calibrated negative log-likelihood):
\[
\operatorname{cNLL}(\bsf, \calD, \tau^\ast) =
- \sum_{(\bsx,y)\in\calD} y \log{\bsp_{\bsf}^{(y)}(\bsx;\tau^\ast)}
\]
\item cECE (calibrated expected calibration error):
\[
\operatorname{cECE}(\bsf, \calD, L, \tau^\ast) = \sum_{l=1}^{L} \frac{|\calB_l|}{|\calD|} \left\lvert
    \operatorname{ACC(\bsf, \calB_l)} - \sum_{(\bsx,\cdot)\in\calB_l} \frac{\max_{k}\bsp_{\bsf}^{(k)}(\bsx;\tau^\ast)}{|\calB_l|}
\right\rvert.
\]
\end{itemize}

\subsection{Ensemble Distillation Methods}\label{app:subsec:distillation_methods}

\paragraph{Ensemble distillation (KD)}
Let $\{\calT_1,...,\calT_M\}$ be a set of pre-trained teachers, and $\calS_\btheta$  be a student.
In practice, we often consider the cross-entropy loss to ground-truth labels during training in addition to the KD loss defined in~\cref{eq:ekd}.
For a given image $\bsx$ and a corresponding one-hot class label $\bsy$, the loss for the KD with the cross-entropy term is defined as
\[\label{eq:ekd_extended}
(1-\alpha) \calH\left[ \bsy,\bsp_{\calS_\btheta}(\bsx;\tau) \right]
+ \alpha \tau^2 \frac{1}{M} \sum_{m=1}^{M} \calH\left[
    \bsp_{\bsp_{\calT_m}}(\bsx;\tau), \bsp_{\calS_\btheta}(\bsx;\tau)
\right].
\]
Here, we have \emph{two hyperparameters}: (1) $\tau$ smooth output probabilities via temperature scaling, and (2) $\alpha$ adjusts the balance between two cross-entropy losses.
We use $(\alpha,\tau)=(1.0, 4.0)$ for experiments on CIFAR-10/100, $(\alpha,\tau)=(0.9, 20.0)$ for experiments on TinyImageNet, and $(\alpha,\tau)=(1.0, 1.0)$ for experiments on ImageNet-1k. 

\paragraph{Adaptive ensemble knowledge distillation (AE-KD)}
\citet{du2020agree} argued that the vanilla ensemble distillation with~\cref{eq:ekd} produces the learning signal determined by the most dominant teacher.
To resolve the issue, they propose Adaptive Ensemble Knowledge Distillation (AE-KD) that optimizes weighting coefficients $\{\omega_m\}_{m=1}^{M}$ of the weighted ensemble distillation loss,
\[
\sum_{m=1}^{M} \omega_m \calH \left[
    \bsp_{\calT_m}(\bsx;\tau), \bsp_{\calS_\btheta}(\bsx;\tau)
\right],
\]
by solving the following optimization problem:
\[
\min_{\omega_1,...,\omega_M} &\quad \frac{1}{2\tau^2} \norm{
    \bsp_{\calS_\btheta}(\bsx) - \sum_{m=1}^{M} \omega_m \bsp_{\calT_m}(\bsx)
}_2^2 \\
\text{subject to} &\quad \sum_{m=1}^{M} \omega_m = 1, \quad 0 \leq \omega_m \leq C \;\;\text{for}\;\; m=1,...,M.
\]
It introduces an \emph{additional hyperparameter}: $C \in [1/M, 1]$ controls the \textit{tolerance of disagreement} among teachers, i.e., with the decrease of $C$, more tolerance of disagreement among the gradients is allowed.
Note that the vanilla ensemble distillation is a specialization of AE-KD when $C=1/M$.
Throughout all experiments, we use $C=0.6$.

\paragraph{Ensemble distribution distillation with Proxy-Dirichlet distribution (Proxy-EnD$^2$)}
\citet{ryabinin2021scaling} introduce a Proxy-Dirichlet target having the density of
\[
\bsq_{\calT}(\bsp|\bsx) = \frac{
    \Gamma\left(\sum_{j=1}^{K}\beta_j(\bsx)\right)
}{
    \prod_{i=1}^{K}\Gamma\left(\beta_i(\bsx)\right)
} \prod_{k=1}^{K} \left( \bsp^{(k)} \right)^{\beta_k(\bsx)-1},
\]
where the concentration parameters are approximated from output probabilities of the teachers,
\[
\beta_k(\bsx) \gets \bsp_{\calT}^{(k)}(\bsx) \frac{
    (K-1)/2
}{
    \sum_{j=1}^{K} \left[
        \bsp_{\calT}^{(j)}(\bsx) \left(
            \log{\bsp_{\calT}^{(j)}(\bsx)} - \frac{1}{M}\sum_{m=1}^{M}\log{\bsp_{\calT_m}^{(j)}(\bsx)}
        \right)
    \right]
}.
\]
Here, $\bsp_{\calT}$ denotes the mean prediction of the teachers, i.e., $\bsp_{\calT}^{(k)}(\bsx) = \frac{1}{M}\sum_{m=1}^{M}\bsp_{\calT_m}^{(k)}(\bsx)$.
This \textit{ensemble distribution} is distilled into the student network $\calS_\btheta$ which represents the density over $(K-1)$-simplex,
\[
\bsq_{\calS_\btheta}(\bsp|\bsx) = \frac{
    \Gamma\left(\sum_{j=1}^{K}\calS_\btheta^{(j)}(\bsx)\right)
}{
    \prod_{i=1}^{K}\Gamma\left(\calS_\btheta^{(i)}(\bsx)\right)
} \prod_{k=1}^{K} \left( \bsp^{(k)} \right)^{\calS_\btheta^{(k)}(\bsx)-1},
\]
where the student network represents the concentration parameters to model the Dirichlet distribution, i.e., the $k^\text{th}$ concentration parameter is $e^{\calS_\btheta^{(k)}(\bsx)}$.
More precisely, they suggest to minimize the reverse KL divergence,
\[
\kld\left(
    \bsq_{\calS_\btheta}(\bsp|\bsx)\;||\;\bsq_\calT(\bsp|\bsx)
\right)
& = \log{\Gamma\left(\sum_{k=1}^{K}e^{\calS_\btheta^{(k)}(\bsx)}\right)} - \sum_{k=1}^{K} \log{\Gamma\left(e^{\calS_\btheta^{(k)}}(\bsx)\right)} \nonumber\\
&\quad\quad + \sum_{k=1}^{K} \log{\Gamma\left(\beta_k(\bsx)\right)} - \log{\Gamma\left(\sum_{k=1}^{K}\beta_k(\bsx)\right)} \nonumber\\
&\quad\quad + \sum_{k=1}^{K} \left( e^{\calS_\btheta^{(k)}(\bsx)} - \beta_k(\bsx) \right) \left( \digamma\left(e^{\calS_\btheta^{(k)}(\bsx)}\right) - \digamma\left(\beta_k(\bsx)\right)\right), \label{eq:proxy_end2_loss}
\]
where $\digamma$ denotes the digamma function.
Throughout experiments, we also follow the practical suggestions: (1) we add one to the concentration parameters, i.e., $e^{\calS_\btheta^{(k)}(\bsx)} \gets e^{\calS_\btheta^{(k)}(\bsx)}+1$ and $\beta_k(\bsx) \gets \beta_k(\bsx) + 1$, and (2) we minimize the loss \cref{eq:proxy_end2_loss} divided by $\sum_{k=1}^{K}\beta_k(\bsx)$ during optimization.

\paragraph{Perturbation strategies}
\citet{srinivas2018jacobian} show that the KD procedure on inputs perturbed by small noise implicitly encourages matching the Jacobians of a teacher and a student.
One can use an isotropic Gaussian noise,
\[
\tilde{\bsx} \gets \bsx + \gamma\bsz,
\]
where $\bsz \sim \calN(\boldsymbol{0},\operatorname{diag}(\boldsymbol{1}))$.
Here, we stay consistent with the most na\"ive choice for the step size, that is, $\gamma=1/255$.
Throughout experiments, we adjust all perturbations to have the same size as the expected size of this Gaussian noise.
Specifically, for the Output Diversified Sampling~\citep[ODS;][]{tashiro2020diversity} perturbation proposed by \citet{nam2021diversity},
\[
\tilde{\bsx} \gets \bsx + \gamma\frac{\nabla_{\bsx}\bsw\tr\bsp_{\calT_m}(\bsx;\tau)}{\norm{\nabla_{\bsx}\bsw\tr\bsp_{\calT_m}(\bsx;\tau)}_2},\quad\text{where } \bsw\sim\calU([-1,1])^K,
\]
we use the step size of $\eta=\sqrt{D}/255$, where $D$ denotes the input dimension, i.e., $\bsx\in[0,255]^{D}$.
Likewise, for our perturbation strategy proposed in~\cref{main:sec:perturbation},
\[
\tilde{\bsx} \gets \bsx + \gamma\frac{\nabla_{\bsx}(\operatorname{TDiv}(\bsx) - \operatorname{SDiv}(\bsx))}{\norm{\nabla_{\bsx}(\operatorname{TDiv}(\bsx) - \operatorname{SDiv}(\bsx))}_2},
\]
we use the same step size of $\eta=\sqrt{D}/255$.

\end{document}